\documentclass[lettersize,journal]{IEEEtran}
\usepackage{amsmath,amsfonts}
\usepackage{algorithmic}
\usepackage{algorithm}
\usepackage{array}
\usepackage[caption=false,font=normalsize,labelfont=sf,textfont=sf]{subfig}
\usepackage{textcomp}
\usepackage{stfloats}
\usepackage{url}
\usepackage{verbatim}
\usepackage{graphicx}
\usepackage{cite}
\usepackage{cite}
\usepackage{bm}
\usepackage{amssymb}
\usepackage{amsmath}
\usepackage{array}   %对表列和表格线的设置需要用到array宏包
\usepackage{tabularx}
\usepackage{multirow}
\usepackage{orcidlink}
\usepackage{tikz}
\usepackage{amsmath,amsfonts}
\usepackage{algorithmic}
\usepackage{algorithm}
\usepackage{array}
\usepackage[caption=false,font=normalsize,labelfont=sf,textfont=sf]{subfig}
\usepackage{textcomp}
\usepackage{stfloats}
\usepackage{url}
\usepackage{verbatim}
\usepackage{graphicx}
\usepackage{multirow}
\usepackage{bm}
\usepackage{amssymb}
\usepackage{amsmath}
\usepackage{array}   %对表列和表格线的设置需要用到array宏包
\usepackage{tabularx}
\usepackage{multirow}
\usepackage{amsmath, amsfonts, amsthm, stackrel, amssymb}   % 用于插入公式
\usepackage{marvosym}

\usepackage{algorithm}
\usepackage{algorithmic}
\usepackage{amsmath, amsfonts, amsthm, stackrel, amssymb} 
\usepackage{colortbl}  %彩色表格需要加载的宏包
\usepackage{xcolor}
\usepackage{array}   %对表列和表格线的设置需要用到array宏包
\usepackage{tabularx}
\usepackage{multirow} 
\usepackage{xcolor}

\usepackage{xcolor,colortbl}
\definecolor{gary}{RGB}{0,120,120}
\definecolor{blue}{rgb}{0.21,0.49,0.74}
\usepackage{bm}
\definecolor{top1}{RGB}{255,179,179}
\definecolor{top2}{RGB}{255,217,179}
\definecolor{top3}{RGB}{255,255,179}
\usepackage[switch]{lineno}
\usepackage{amsmath, amsfonts, amsthm, stackrel, amssymb}   % 用于插入公式
\begin{document}
\title{Hi-LSplat: Hierarchical 3D Language Gaussian Splatting}

\author{Chenlu Zhan, Yufei Zhang, Gaoang Wang ~\IEEEmembership{Member,~IEEE}, Hongwei Wang~\IEEEmembership{Senior Member,~IEEE}
        % <-this % stops a space
\thanks{This work was supported in part by Zhejiang Provincial Natural Science Foundation of China (LDT23F02023F02) and  the National Natural Science Foundation of China (No.62106219). \emph{Corresponding Authors: Hongwei Wang and Gaoang Wang.}}% <-this % stops a space
\thanks{Chenlu Zhan is with the College of Computer Science and Technology, Zhejiang University, Zhejiang, China (chenlu.22@intl.zju.edu.cn)}
\thanks{Yufei Zhang is with the College of Biomedical Engineering and Instrument Science, Zhejiang University, Zhejiang, China (yufei1.23@intl.zju.edu.cn)}
\thanks{Gaoang Wang, and Hongwei Wang are with Zhejiang University-University of Illinois Urbana-Champaign Institute, Zhejiang University, Haining, China. (gaoangwang@intl.zju.edu.cn, hongweiwang@intl.zju.edu.cn) 
}
}

\maketitle

\begin{abstract}
Modeling 3D language fields with Gaussian Splatting for open-ended language queries has recently garnered increasing attention. 
However, recent 3DGS-based models leverage view-dependent 2D foundation models to refine 3D semantics but lack a unified 3D representation, leading to view inconsistencies. Additionally, inherent open-vocabulary challenges cause inconsistencies in object and relational descriptions, impeding hierarchical semantic understanding.
In this paper, we propose \textbf{Hi-LSplat}, a view-consistent \textbf{Hi}erarchical \textbf{L}anguage Gaussian \textbf{Splat}ting work for 3D open-vocabulary querying. 
To achieve view-consistent 3D hierarchical semantics, we first lift 2D features to 3D features by constructing a 3D hierarchical semantic tree with layered instance clustering, which addresses the view inconsistency issue caused by 2D semantic features.
Besides, we introduce instance-wise and part-wise contrastive losses to capture all-sided hierarchical semantic representations.
Notably, we construct two hierarchical semantic datasets to better assess the model's ability to distinguish different semantic levels.
Extensive experiments highlight our method's superiority in 3D open-vocabulary segmentation and localization. 
Its strong performance on hierarchical semantic datasets underscores its ability to capture complex hierarchical semantics within 3D scenes.
%Its superior performance on hierarchical semantic datasets highlights its effectiveness in capturing complex hierarchical semantics within 3D scenes.
\end{abstract}

\section{Introduction}
\label{sec:intro}

3D open-vocabulary query enhance human interaction with 3D environments~\cite{simvqaCascante-Bonilla_2022_CVPR,Azuma_2022_CVPRscanqa},  benefiting 3D semantic segmentation ~\cite{ying2024omniseg3d,qin2024langsplat,Yin_2024_CVPRSAI3D,ConvolutionsDieHard}, virtual reality~\cite{3DOVS}, and robotic navigation ~\cite{robot} applications. Recent studies~\cite{qin2024langsplat,wu2024opengaussian} have emphasized modeling 3D language fields to support open-vocabulary queries, underscoring the importance of consistent, all-sided hierarchical semantics in 3D scenes.
\begin{figure}[!htp]
\centering
\includegraphics[width=1\linewidth]{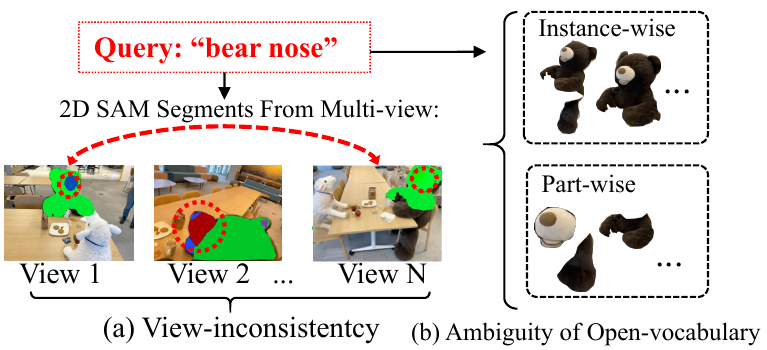}
\caption{Limitations of (a) Inconsistent 2D segmentation across multi-views (2) Ambiguity of open-vocabulary.}
%\caption{Comparison of the (a) 2D-based association works and (b) Our 3D point-based hierarchical cluster method.}
\label{compare_2d3d}
\end{figure}

Recent works~\cite{chou2024gsnerf,zuo2024fmgs,wu2024opengaussian,chen2024pgsr} leverage efficient 3D Gaussian Splatting~\cite{kerbl20233gs} to embed language attributes into 3D Gaussian representations~\cite{li2025aligning,yu2024jimr,lin2025delving}.
However, these methods rely on 2D techniques like CLIP~\cite{radford2021clip} and SAM~\cite{Kirillov_2023_ICCVsam} to project language properties onto images, maintaining 3D scene consistency through multi-view 2D features, which introduces significant limitations:
%However, these methods primarily rely on 2D foundational models such as CLIP and SAM to map semantic attributes onto images, maintaining 3D scene consistency through 2D features. This approach of lifting 2D technical to optimize 3D scene understanding has significant limitations:
%(1) Inconsistent hierarchical semantics. The inherent flatness of 2D features neglects occluded and underside areas in the training data, leading to biases in 3D scene comprehension. (2) Lack of hierarchical semantic correlation. Models based on 2D pixel-level features fail to align with downstream tasks that require 3D point-level scene understanding, interaction, segmentation, and localization. 
{(1) View-inconsistentcy by 2D pixel-aligned semantic feature.} Most works, such as LangSplat~\cite{qin2024langsplat}, extract 2D semantic features using view-dependent 2D foundation models but lack a unified 3D point-alined representation for scene understanding, leading to view inconsistencies, as shown in Fig.~\ref{compare_2d3d} (a). This results in noisy segmentation and distorted semantics, especially for hierarchical objects (e.g., eaves and roofs).
%Most works like LangSplat~\cite{qin2024langsplat} adopt 2D pixel-aligned optimization methods which maps multi-view images to 3D suffer from {view-inconsistency}, as shown in Fig.~\ref{compare_2d3d} (a), causing noisy segmentation and distorted semantics, particularly for {hierarchical objects} (e.g., eaves vs. roof).  %Even with 2D masks, single-object segmentation and occlusion instability hinder accurate hierarchical semantics, thus constraining the semantic query of 3D scenes. 
%The disparity in 2D segmentation across viewpoints leads to multi-view inconsistencies and substantial ambiguity in handling complex scenes with hierarchical semantic structures, as illustrated in Fig.~\ref{compare_2d3d} (a), thus constraining the semantic query of 3D scenes. 
%2D pixel optimization methods mapping multi-view images to 3D suffer from \textbf{view-inconsistency}, causing noisy segmentation and distorted semantics, particularly for \textbf{hierarchical objects} (e.g., eaves vs. roof).  Even with 2D masks, single-object segmentation and occlusion instability hinder accurate hierarchical semantics.
%The disparity in 2D segmentation across different viewpoints results in multi-view inconsistency, causing significant ambiguity when dealing with complex scenes and inherently hierarchical semantic structures, thereby limiting the ability to comprehend 3D scenes.
{ (2) Lack of hierarchical semantic distinction in open-vocabulary queries.} The lack of hierarchical semantics is an inherent issue in open-vocabulary queries,  as Fig.~\ref{compare_2d3d} (b) shows. 
 Relevant works like LangSplat~\cite{qin2024langsplat} reveal the ambiguity of open-vocabulary settings and inconsistencies in describing object-level and hierarchical semantic relationships, with semantic features constrained to basic categories or local geometric properties, particularly for {hierarchical objects} (e.g., eaves vs. roof).
 %Relevant works like LangSplat~\cite{qin2024langsplat} show inconsistencies in object and relational descriptions, with semantic features limited to basic categories or local geometry, especially for \textbf{hierarchical objects} (e.g., eaves vs. roof). 
 Additionally, in 3DGS, a single Gaussian point representing multiple pixels introduces feature similarity, further blurring hierarchical semantics.

In this paper, we propose a novel Hierarchical Language Gaussian Splatting method, namely Hi-LSplat, for 3D open-vocabulary querying. While recent advancements such as LangSplats~\cite{qin2024langsplat} and OpenGaussian~\cite{wu2024opengaussian} have made strides in 3D scene understanding, our model distinguishes itself through two fundamental innovations: a dedicated focus on hierarchical 3D semantics and the curation of two specialized datasets. This represents a departure from prior works, which primarily address basic query tasks. To validate our approach, we contribute two novel hierarchical semantic datasets that underscore the superiority of our model in capturing multi-level semantic relationships.
Secondly, we introduce a 3D hierarchical semantic tree to enforce view-consistent feature representation, directly addressing the cross-view inconsistency issues plaguing methods like LangSplats~\cite{qin2024langsplat}. In contrast, OpenGaussian~\cite{wu2024opengaussian} relies on view-agnostic SAM boolean masks for point-level open-vocabulary understanding. Our method transcends this by constructing a semantic hierarchy from view-consistent features, enabling deeper comprehension of nested semantic structures within 3D scenes.
Finally, we design instance and part contrastive learning mechanisms to model hierarchical semantic associations, a capability absent in existing frameworks that focus solely on 2D planar relationships. This allows our model to reason across spatial scales and semantic levels, establishing a new paradigm for fine-grained 3D scene understanding.

%The key difference between our and is our focus on 3D \textbf{hierarchical semantics and the creation of {two datasets}}, while others address only basic queries. (\textcolor{red}{Fig. 4-9, Table 4, Appendix A, D}). (2) We introduce a 3D hierarchical semantic tree for \textbf{view-consistent features}, unlike others with cross-view inconsistency (\textcolor{red}{Fig. 8-9, Appendix E}). (3) We design instance and part contrastive learning for \textbf{hierarchical semantic associations}, whereas others focus only on 2D planar relationships (\textcolor{red}{Fig. 2, 4, 8}).

Specifically, the key to achieving view-consistent semantic representations is optimizing independently generated 3D instance features rather than relying on view-inconsistent 2D features. We derive 3D instance features by applying view-independent SAM boolean masks (instead of high-dimensional mask features) to the feature map, computing the mean feature for each binary mask region. A point-optimized clustering loss then aligns each 3D instance feature with its mean, mitigating viewpoint inconsistencies without requiring multi-view associated high-dimensional 2D masks.
Additionally, we extract 3D clustered features at different semantic levels to construct a 3D hierarchical cluster tree. We then employ instance-level contrastive learning to encode hierarchical semantic similarities and part-level learning to capture internal hierarchical relationships. 
Notably, we reconstruct two hierarchical semantic datasets to better evaluate the model's capability in distinguishing hierarchical semantics.
%To avoid the inconsistencies inherent in elevating 2D image understanding to 3D space, we first use SAM's binary masks and instance clustering to train distinctive yet 3D-consistent point-level instance features. To capture the complex multi-object and hierarchical semantic features of 3D scenes, we design a multi-level hierarchical semantic relation tree, utilizing masks for instances, parts, and sub-parts obtained from SAM. These masks are assigned hierarchical relationships based on coverage IoU. Our method employs instance-wise contrastive learning loss to represent semantic structure relationships between instances and part-wise contrastive learning to precisely capture the subordinate relationships between instances and their corresponding parts and sub-parts, thus learning both global and local semantic representations of 3D scenes. 
%Experimental results on open-vocabulary 3D object localization and semantic segmentation tasks validate our method's significant advantages in 3D scene understanding and its effective comprehension of instance-wise and part-wise complex semantic information.
%We conducted open-vocabulary 3D object localization and semantic segmentation tasks on three datasets. Extensive experimental results validate the significant advantages of our method in 3D scene understanding and its effective comprehension of complex semantic information at both instance and part levels.
\begin{figure*}[!htp]
\centering
\includegraphics[width=1\linewidth]
{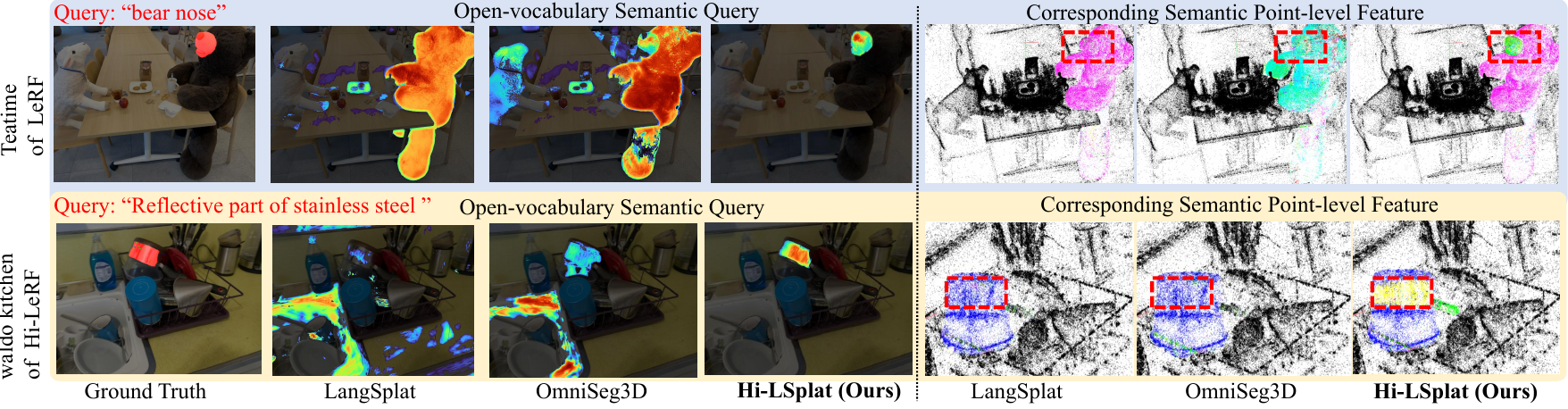}
\caption{Comparison of 3D open-vocabulary semantic query and semantic feature between our model and SOTA hierarchical semantic models. We highlighted the semantic features most relevant to the query, with different colors representing distinct features. Hi-LSplat excels at capturing precise 3D hierarchical semantics and accurately segmenting hierarchical features.}
%Comparison of complex semantic querying between our FreeQ-Graph, designed for free-form querying, and previous SOTA 3D open-vocabulary semantic segmentation methods.}}
\label{fig:0}
\end{figure*}
We conduct semantic segmentation and localization tasks on 8 datasets,  including 6 public datasets
and 2 constructed hierarchical datasets for semantic and instance segmentation, and localization tasks.
, demonstrating our significant advantage in capturing 3D consistent and hierarchical semantics.
%, while its superior performance on hierarchical semantic datasets underscores its efficacy in capturing 3D hierarchical semantics.
Our contributions are summarized as follows:
\begin{itemize}
%\item We propose Hi-LSplat, a hierarchical semantic point-optimized work based on language Gaussian Splatting for 3D open-vocabulary querying.

\item We propose a view-consistent 3D hierarchical semantic-guided Language Gaussian field, utilizing a hierarchical tree with layered point-optimized instance clustering for 3D view-dependent semantic features.

\item We propose instance-wise and part-wise contrastive learning to represent external and internal hierarchical semantic relations in open-vocabulary queries.
%the semantic structure and subordinate relationships of complex scenes.

\item We reconstruct two hierarchical semantic datasets for improved evaluation. Experiments on 8 datasets show that our method outperforms others in achieving 3D consistent and hierarchical semantics, with improvements of 34.14 and 8.0 mIoU on ScanNet and LERF datasets.

\end{itemize}

\section{Related Works}
\label{sec:formatting}
\subsection{3D Open-vocabulary Query}

Recent 3D scene open-vocabulary query works have benefited from advances in 2D segmentation techniques, such as SAM~\cite{Kirillov_2023_ICCVsam} and its variants~\cite{yang2023sam3dsegment3dscenes,cen2024segment3dgaussians,choi2024clickgaussianinteractivesegmentation3d}.  
They integrate semantic features from 2D models like CLIP~\cite{radford2021clip} and DINO~\cite{zhang2022dinodetrimproveddenoising} into scene representations by NeRF~\cite{mildenhall2021nerf} and 3D-GS~\cite{kerbl20233gs} to improve 3D scene understanding~\cite{xue2023uliplearningunifiedrepresentation,liao2024clipgsclipinformedgaussiansplatting,ji2024fastlgsspeedinglanguageembedded}, segmentation~\cite{Ding_2023_CVPRPLA,takmaz2023openmask3dopenvocabulary3dinstance,Peng_2023_CVPRopenscene,guo2024semanticgaussiansopenvocabularyscene}, and editing~\cite{Wang_2024_CVPRGAUSSIANEDITOR,Editanything,liao2024clipgsclipinformedgaussiansplatting}.
Despite their focus on adapting 2D techniques for 3D scene semantic representations through cross-view consistency, these methods~\cite{shi2024LEGaussians, kerr2023lerf, qin2024langsplat, zhou2024feature3dgs, silva2024contrastivegaussianclusteringweakly, ye2024gaussiangroupingsegmentedit,yang2024tuningfreeuniversallysupervisedsemanticsegmentation} still struggle with inherent inconsistencies and biases.
Gaussian Grouping~\cite{ye2024gaussiangroupingsegmentedit} uses SAM-extracted masks to train 2D view consistency for reconstructing and segmenting open 3D scenes. LEGaussians~\cite{shi2024LEGaussians} leverages dense pixel features from CLIP~\cite{radford2021clip} and DINO~\cite{zhang2022dinodetrimproveddenoising}, introducing semantic attributes for each Gaussian to constrain the rendered semantic map.
Recent works, such as OpenGaussian~\cite{wu2024opengaussian} and CGC~\cite{silva2024contrastivegaussianclusteringweakly}, focus on learning 3D consistent point-level instances. However, these instance-level clustering approaches~\cite{Liang_2023_CVPRseg,Shin_2024_CVPRsphine} fail to represent the complex semantic relationships in complex 3D scenes.
Our method not only directly learns 3D view-consistent semantic features but also builds a 3D semantic hierarchy tree, allowing for a precise and comprehensive hierarchical understanding of 3D scenes.

\begin{figure*}[h]
\centering
\includegraphics[width=1\linewidth]{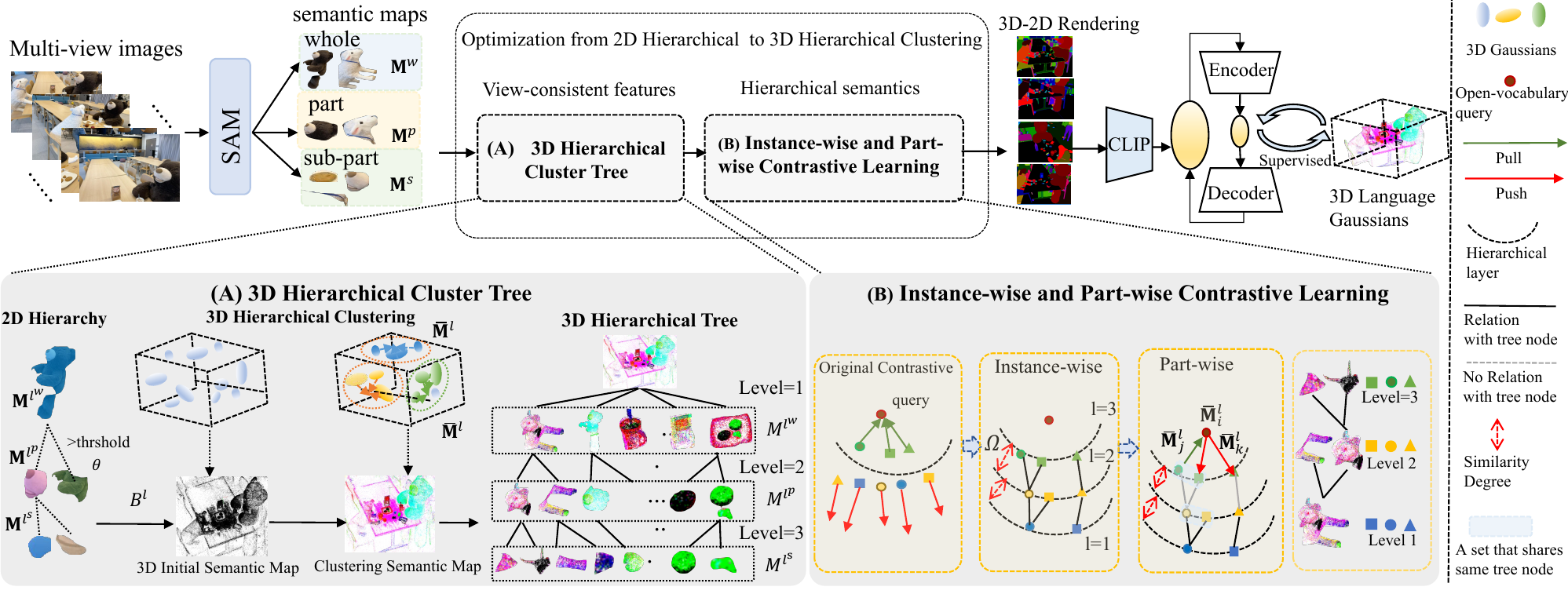}
\caption{{The network structure of Hi-LSplat.} We propose a view-consistent hierarchical language Gaussian Splatting work for 3D open-vocabulary querying with (A) 3D Hierarchical semantic cluster tree for view-consistent and hierarchical semantic features, and (B) Instance-wise and part-wise contrastive learning for external and internal semantic correlation.}
\label{main}
\end{figure*}
\subsection{Hierarchical Scene Representation}
Hierarchical representations aid in analyzing the geometric-semantic relationships among the complex scenes~\cite{cao2024sohesselfsupervisedopenworldhierarchical}. Existing hierarchical semantic methods~\cite{bspnetChen_2020_CVPR,hiChen_2021_ICCV,hi3Qiu_Liu_Chen_King_2024,hi2NEURIPS2023_43663f64,hi4Xu_2023_ICCV,Kang_2024_CVPRHierarchical,Mo_2019_CVPRpartnet} predominantly focus on semantic analysis in 2D planes and for single objects within specific categories.
Current methods~\cite{qin2024langsplat,ying2024omniseg3d, Kim_2024_CVPRgrop,Xu_2023_ICCVhi2,maskgroup} focus on deriving planar-level hierarchical semantics by leveraging scale variations across different object categories and 2D masks generated from segmentation models such as SAM~\cite{Kirillov_2023_ICCVsam}.
LangSplat~\cite{qin2024langsplat} utilizes SAM to extract, align, and compress 2D hierarchical masks, embedding them into 3DGS~\cite{kerbl20233gs} for enhanced semantics. GARField~\cite{Kim_2024_CVPRgrop} decomposes 3D scenes into semantic groups based on physical scales derived from images. OmniSeg3D~\cite{ying2024omniseg3d} uses a class-agnostic 2D segmentation hierarchy to model multi-level pixel relationships and generate 3D feature fields. VCH~\cite{he2024viewconsistenthierarchical3dsegmentation} introduces a novel feature space by applying varying thresholds to feature distances, enabling segmentation across different scales.
However, these methods depend on indirect 2D pixel-level hierarchies, resulting in mismatches with 3D scene semantics and overlooking the intricate hierarchical of 3D scenes. For ours, we capture all-sided 3D point-level semantics, representing both instance-wise global structures and local part-wise relationships.

\section{Method}
In this section, we propose Hi-LSplat,  a view-consistent hierarchical language Gaussian Splatting work for 3D open-vocabulary querying with (A) Hierarchical scene semantic tree with 3D hierarchical cluster for view-consistent semantic features, and (B) Instance-wise and part-wise contrastive learning for global and local semantic correlation,  as illustrated in Fig.~\ref{main}.
\subsection{Preliminary on Language Gaussian Splatting}
\label{Preliminary on Semantic feature with Gaussian Splatting}
%The representation of 3D Gaussians as point clouds explicitly depicts 3D scenes. 
3D Gaussian Splatting~\cite{kerbl20233gs} explicitly depict the 3D scene through numerous Gaussian points.
%Each Gaussian point has attributes like a covariance matrix $\Sigma$ and a centroid $\chi$, with mean feature $G(X) = e^{ - \frac{1}{2}{\chi ^T}{\Sigma ^{ - 1}}\chi }$.
%For differentiable optimization, $\Sigma$ is composed of a scaling matrix $M$ and a rotation matrix $R$: $\Sigma = RM{M^T}{R^T}$. 3D Gaussians use differential splitting within camera planes for novel view rendering. The covariance matrix $\Sigma'$ in camera coordinates is calculated via the transform matrix $W$ and the Jacobian matrix $J$ from the affine approximation of the projective transformation: ${\Sigma} ^{'} = JW\Sigma W^T{J^T}$. 
%Each 3D Gaussian has attributes including SH coefficients ${\rm C} \in \mathbb{R}^k$, opacity $\alpha \in \mathbb{R}$, rotation $r \in \mathbb{R}^4$, scaling $s \in \mathbb{R}^3$, and position $\chi \in \mathbb{R}^3$. 
Each 3D Gaussian has attributes including SH coefficients ${\rm C}$, opacity $\alpha$, rotation $r$, scaling $s$, and position $x$.
%For 2D rendering, 3D Gaussian Splatting arranges $N$ Gaussians contributing to a pixel, combining overlapping Gaussians: $C = \sum\limits_{i \in N} {{c_i}} {\alpha _i}\prod\limits_{j = 1}^{i - 1} {(1 - } {\alpha _i})$
%of the point from the 3D Gaussian $G$ with covariance $\Sigma$, multiplied by adjustable opacity and SH color coefficients.
For 3D semantic features, following previous works~\cite{qin2024langsplat,wu2024opengaussian}, we enhance each 3D Gaussian with a learnable low-dimensional 3D semantic feature $\mathbf{f}\in \mathbb{R}^d$ to represent language attributes. %Following previous works~\cite{qin2024langsplat,wu2024opengaussian},
For any given training views, we similarly follow the Gaussian Splatting process, using alpha-blending to render the 3D instance feature $\mathbf{f}\in \mathbb{R}^d$ into a semantic feature map $\mathbf{M} \in \mathbb{R}^{d \times H \times W}$:
\begin{equation}
\label{M}
    \mathbf{M} = \sum\limits_{i \in N} {{\mathbf{f}_i}} {\alpha _i}\prod\limits_{j = 1}^{i - 1} {(1 - } {\alpha _i})
\end{equation}
where ${\alpha_i}$ is the density and color of the Gaussian point, $d$ is the dimension of CLIP embedding after an autoencoder. 
%This enhancement remains consistent across all viewing directions. 
%Therefore, we extract components directly from the Gaussian function without calculating spherical harmonic coefficients. 

%\subsubsection{Gaussian Splatting}
%\subsubsection{}

\subsection{3D Hierarchical Semantic Tree}
\label{Hierarchical scene semantic tree}

%To effectively capture the semantic hierarchy of 3D scenes, we constructed a 3D hierarchical tree in two main steps: 1) We derive the initial 2D semantic hierarchy by analyzing the overlap relationships among segmentation maps at three different levels, generated from SAM for each input image.
%Using SAM, we provided each input image with segmentation maps at three different scales: \emph{whole, part, and subpart}.  We built the 2D hierarchical tree based on the coverage relationships among these three levels of masks.
To effectively capture the semantic hierarchy of 3D scenes, we construct a 3D hierarchical tree in two main steps: 1) We derive an initial 2D semantic hierarchy by analyzing the overlap between the three segmentation levels generated by SAM for each input image.
2) We train and cluster point-level features based on the initial 2D hierarchical relationships, forming a consistent 3D semantic hierarchy tree.

\noindent \textbf{Initialize 2D Semantic Feature Maps.}
Specifically, following LangSplat~\cite{qin2024langsplat}, we first utilized SAM~\cite{Kirillov_2023_ICCVsam} with a $32\times 32$ point-prompt grid to segment the same input image view into different semantic levels of masks: \emph{whole, part, and subpart}. We then generate three distinct hierarchical semantic feature maps: $\mathbf{M}^w$, $\mathbf{M}^p$, and $\mathbf{M}^s$ through Eqn.~\ref{M}, representing the entire image by combining predicted IoU, stability scores, and overlap rates among the three mask levels.
Extracting 2D semantic targets alone fails to capture view-consistent hierarchical semantics in complex 3D scenes. Hence, we model the inclusion relationships among these features to enable the model to grasp the semantic structure between objects and their components.

We incorporated hierarchical associations into the masks by analyzing overlaps across the three mask types. Following OpenGaussian~\cite{wu2024opengaussian}, we determine the hierarchical relationship by checking if the overlap between two masks exceeds a threshold, thus establishing distinct semantic levels, as the left side of Fig.~\ref{main} (A) shown.
First, we set a coverage threshold  $\theta$. 
If the following three conditions are met: 1) Over $\theta$ of pixels in mask $A$ are also in mask $B$. 2) Less than $\theta$ of pixels in mask $B$ are in mask $A$. 3) Mask $B$ is the smallest mask that meets the first two conditions. Mask $A$ is considered covered by mask $B$, meaning $A$ is a child node of $B$. 
We apply this process to the three semantic level maps $\mathbf{M}^w$, $\mathbf{M}^p$, and $\mathbf{M}^s$, generating hierarchical representations for different semantic features, denoted as $\mathbf{M}^{l^w}$, $\mathbf{M}^{l^p}$, and $\mathbf{M}^{l^s}$.
Each layer's semantic masks are labeled with its tree hierarchy level 
$l$, specifically: $l^w=1$, $l^p=2$, and $l^s=3$.
%Details can be found in the next Section.
%: whole, part, and subpart. 
%Finally, we consider the root masks of all tree structures as ``whole" objects, their direct child objects as ``part" objects, and all remaining descendants as ``subpart" objects.
%Each of these three levels of objects is labeled with their tree hierarchy level $l$, which are:  $l^w=1$, $l^p=2$, and $l^s=3$.

\noindent \textbf{3D Hierarchical Semantic Clustering.}
\label{3dtree}
%Unlike previous methods~\cite{qin2024langsplat,wu2024opengaussian,kerr2023lerf} that primarily use 2D semantic understanding technics to align 2D and 3D semantic features and then enhance 2D features for view-consistency tasks in 3D scene understanding, 
%our model does not rely on view-consistent 2D approaches. 
We train and cluster point-level features based on initial 2D hierarchical relationships, constructing a view-consistent 3D semantic hierarchy tree.
Unlike previous methods~\cite{qin2024langsplat,wu2024opengaussian,kerr2023lerf}, which suffer from 2D multi-view inconsistency, our approach ensures view consistency by leveraging view-independent SAM Boolean masks, rather than high-dimensional mask features for 3D point-based clustering. This strategy facilitates the optimization of view-consistent 3D semantic features.

Notably, for each viewpoint, we generate the corresponding 2D masks and categorize them into three distinct semantic levels based on the threshold $\theta$, which are then used for subsequent instance clustering. Since the 3D semantic hierarchy tree is constructed through 3D instance-level clustering, enforcing cross-view consistency at the 2D level is no longer required.

%Notably, for each viewpoint, we generate corresponding 2D masks and categorize them into three distinct semantic levels based on $\theta$, which will be used for subsequent instance clustering. Since the 3D semantic hierarchy tree is derived from 3D instance clustering, 2D cross-view consistency does not need to be considered.
%Instead, we directly use 3D Gaussian point cloud instance consistency for instance clustering and further integrate the 2D hierarchical tree to obtain accurate semantic hierarchical instances with 3D consistency.

Given any training view, we can obtain three types of hierarchical feature maps $\mathbf{M}^{l^w}$, $\mathbf{M}^{l^p}$, and $\mathbf{M}^{l^s}$ from instance features through alpha blending, along with  %hierarchical tree comprising three types of masks representing different semantic levels and
their respective hierarchical levels. Following OpenGaussian~\cite{wu2024opengaussian}, we first define the average feature $\bar {{{\mathbf{M}}}_i^l}  = \left( {{\bf{B}}_i^l \cdot {{{\mathbf{M}}}^l}} \right)/\sum {{\bf{B}}_i^l}  \in {\mathbb{R}^3}$ within each boolean mask, where 
$\boldsymbol{B}^l_i \in\{0,1\}^{1 \times H \times W}$ represents the $i$-th mask at the 
$l$-th semantic level. 
%Based on 3D Gaussian scene consistency, we aim for the Gaussian-rendered features within the same mask to be close to their average value $\bar {\bf{M}}$, with different levels of masks representing distinct semantic information in the 3D scene. Therefore, we designed a 3D hierarchical clustering loss, defined as follows:
%To learn consistent features in the 3D Gaussian scene, our goal is to make the Gaussian-rendered features within the same mask approach their mean value $\bar {{M}}$. 
To ensure consistency in 3D Gaussians, our goal is for Gaussian-rendered features within the same mask to converge to their mean value $\bar{\mathbf{M}}$. 
Additionally, to capture varying semantic levels of the 3D scene, we designed a point-optimized hierarchical clustering loss defined as follows:
\begin{equation}
   % {{\cal L}_{\rm{h}}} =  \frac{1}{{L}}\sum\limits_{l = 1}^L {\sum\limits_{i = 1}^m {\sum\limits_{h = 1}^H {\sum\limits_{w = 1}^W {{\bf{B}}_{i,h,w}^l} } }  \cdot {{\left\| {{{M}}_{:,h,w}^l - \bar {{{M}}_i^l} } \right\|}^2}} 
   \mathcal{L}_{\mathrm{h}}=\frac{1}{L} \sum_{l=1}^L \sum_{i=1}^m\left\|\mathbf{B}_i^l \cdot\left(\mathbf{M}^l-\bar{\mathbf{M}}_i^l\right)\right\|^2
\end{equation}
where $L$ represents the hierarchical level of masks, $m$ is the total number of hierarchical masks at each level. 
%and $H$ and $W$ are the height and weight of the image view. 
Through hierarchical instance clustering, we can learn view-consistent 3D semantic features with their hierarchical information.
We train and cluster point-level features based on view-independent boolean masks, resulting 3 levels of semantic features forming 3D semantic hierarchy tree.

%These three distinct semantic levels form the proposed three-level 3D semantic feature tree.

\subsection{Instance-wise and Part-wise Contrastive}
%To capture the global and local hierarchical semantic relationships in complex 3D scenes, our method differs from previous approaches that simply separate instances. We designed an instance-wise loss $L_{ins}$ to grasp the relationships between masks within the semantic hierarchical structure and a part-wise loss $L_{part}$ to learn the relationships between masks and their corresponding sub-part similarities.
%To capture global and local hierarchical semantic relationships in complex 3D scenes, our method departs from previous approaches that simply separate different instances. We designed an all-encompassing semantic hierarchy learning approach: 1)An instance-wise loss $L_{ins}$to grasp the external semantic hierarchy relationships between masks.2) A part-wise loss $L_{part}$ to capture the internal hierarchical semantics among features.
To capture both global and local hierarchical semantics in complex 3D scenes, our method moves beyond the conventional approach of merely distinguishing instances. We propose a all-sided semantic hierarchy learning framework: 1) An instance-wise loss $L_{ins}$ to capture external semantic hierarchies between masks; 2) A part-wise loss $L_{part}$ to model internal hierarchical relationships among features.
%Specifically, for the hierarchical instance-wise loss, we aim to increase the distance between the average features of different instances to enhance feature diversity, while also enhancing the similarity of masks at different hierarchical levels based on their level. 
\noindent \textbf{Instance-wise.} 
%Specifically, for the hierarchical instance-wise loss, our objectives are twofold: (1) to increase the distance between different mean features $\bar {{M}}$ to enhance the diversity of distinct features, and (2) to differentiate the feature similarity across hierarchical semantic levels.
The hierarchical instance-wise loss has two objectives: (1) to increase the distance between different mean semantic features $\bar{\mathbf{M}}$ to enhance feature diversity, and (2) to differentiate semantic similarity across hierarchical semantic levels.
Given the average features 
%$\bar {M^{l_i}} $, $\bar {M^{l_j}}$ 
$\bar {\mathbf{M}^{l}_i} $, $\bar {\mathbf{M}^{l}_j}$ 
of two different masks with tree levels $ {l_i} $, $ {l_j}$, their semantic similarity is denoted as 
$| {l_i}  - {l_j}|$. Drawing inspiration from the hierarchical clustering loss in ~\cite{Yan_2021_CVPRmetric}, which approximates similarity through distance ratios in the feature embedding space, we assign varying similarity degrees $\Omega$ to features across different semantic levels. Thus, we can push apart negative masks with varying margins guided by intrinsic similarity levels. The hierarchical instance-wise loss is defined as follows:
\begin{equation}
   \resizebox{0.90\hsize}{!}{$ {{\cal L}_{ins}} =\frac{1}{{N_t}({N_t}-1)} \sum\limits_{i = 1}^{N_t} {\sum\limits_{j = 1,j \ne i}^{N_t} {(\log \frac{1}{{\left\| {\bar {\mathbf{M}_i^l}  - \bar {\mathbf{M}_j^l} } \right\|}}} }  - \log {\Omega ^{|{l_i^l}  -  {l_j^l} |}}{)^2}$}
\end{equation}
where $\Omega>0$ is the hyperparameter to trade off the similarity, ${N_t}$ is the total number of masks in the semantic tree.

%In addition to inter-instance similarity, complex 3D scenes require consideration of semantic relationships between different parts and sub-parts, such as distinguishing between ``bear nose" and ``bear mouth," which belong to the same subclass but have different semantic information. 
\noindent \textbf{Part-wise.} Beyond instance-wise similarity, complex 3D scenes necessitate consideration of the internal semantic relationships among features. For example, distinguishing ``bear nose" and ``bear mouth" which belong to the same subclass but carry distinct semantic information.
Utilizing our constructed 3D hierarchical tree, we first eliminate semantic overlap between two different mean features and emphasize their differences in the loss function. Specifically, when computing the similarity between 
$\bar {\mathbf{M}_i^l} $, $\bar {\mathbf{M}_j^l}$, we subtract their tree node of mean features at the previous hierarchical level ${ \mathbf{\bar M}}^{l - 1}$, resulting in a new similarity score $s_p$, which can be denoted as follows:
\begin{equation}
     {s_p}\left( {{{ \mathbf{\bar M}}^l}_i,{{ \mathbf{\bar M}}^l}_j} \right) = \frac{{{{\left( {{{ \mathbf{\bar M}}^l}_i - {{ \mathbf{\bar M}}^{l - 1}}} \right)}^T}\left( {{{ \mathbf{\bar M}}^l}_j - {{ \mathbf{\bar M}}^{l - 1}}} \right)}}{{\left\| {{{ \mathbf{\bar M}}^l}_j - {{ \mathbf{\bar M}}^{l - 1}}} \right\|\left\| {{{ \mathbf{\bar M}}^l}_j - {{ \mathbf{\bar M}}^{l - 1}}} \right\|}}
\end{equation}

%Based on the new similarity computation method, we design a part-wise loss that compares the representations of shared semantics between the average features of two different masks, rather than merely separating their distinct semantics. These semantics are disentangled with the help of the 3D hierarchical tree and vary according to the relative positions of the category pairs, which can be defined as follows:
Based on the new similarity computation method, we devised a part-wise internal loss function that separates distinct semantic features without comparing representations of their common semantics. These semantics are disentangled with the aid of the 3D hierarchical semantic tree and vary according to the relative hierarchy of semantic features.
Besides, we define ${|\bar S_P^l|}$ $( P \in [1, N_p])$ as a set that shares the same tree node features in the $l$-th hierarchy layer, and ${{ \mathbf{\bar M}}^l}_i$ is the $i$-th subfeature in ${|\bar S_P^l|}$. In each semantic hierarchy $l$ in the 3D semantic hierarchy tree, for each selected average clustering feature ${{ \mathbf{\bar M}}^l}_i$, we respectively take the semantic features ${{ \mathbf{\bar M}}^l}_j$ that share the same tree features as positive samples, while others ${{ \mathbf{\bar M}}^l}_k$ in the same semantic hierarchy but not in the ${|\bar S_P^l|}$ set as negative samples as defined below:
\begin{equation}
  %\resizebox{0.905\hsize}{!}{$  {{\cal L}_{part}} =  - \frac{1}{{L{m}}}\sum\limits_{l = 1}^L {\sum\limits_{i = 1}^{{m}} {\sum\limits_{j = 1}^{\left| {{{\bar M}^l_j}} \right|} {\log } } \frac{{\exp \left( {{s_p}({{\bar M}^l}_i,{{\bar M}^l}_j)/\tau } \right)}}{{\sum\limits_{k = 1}^{{m}} {\exp } \left( {{s_p}({{\bar M}^l}_i,{{\bar M}^l}_k)/\tau } \right)}}} $}
  \resizebox{0.905\hsize}{!}{$ {{\cal L}_{part}} =  - \frac{1}{{L{N_p}}}\sum\limits_{l = 1}^L {\sum\limits_{i = 1}^{{N_l}} {\sum\limits_{j = 1}^{\left| {\bar S_P^l} \right|} {\log } } \frac{{\exp \left( {{s_p}({{\mathbf{\bar M}}^l}_i,{{\mathbf{\bar M}}^l}_j)/\tau } \right)}}{{\sum\limits_{k = 1}^{{N_k}} {\exp } \left( {{s_p}({{\mathbf{\bar M}}^l}_i,{{\mathbf{\bar M}}^l}_k)/\tau } \right)}}}   $}
\end{equation}
where $\tau$ is the temperature parameter. $N_p$ is the number of positive features, $N_k$ is the number of negative features, $N_l$ is the number of mask features in the $l$-th layer in semantic tree. The $L$ is the total level of the hierarchical semantic tree.

\subsection{Training and Inference}
The overall objective function includes hierarchical clustering loss $L_h$, instance-wise $L_{ins}$ and part-wise contrastive learning losses $L_{part}$, re-weighted by parameters $\lambda_1$ and $\lambda_2$:
\begin{equation}
    L = {L_h} + {\lambda _1}{L_{ins}} + {\lambda _2}{L_{part}}
\end{equation}
%It is noteworthy that our method not only performs hierarchical clustering directly on 3D point-level instances to  obtain consistent semantic features and hierarchical information but also captures both global and local relationships within the semantic hierarchical structure and between masks and their corresponding sub-parts. This approach enables a comprehensive and direct learning of the global and local hierarchical semantics of the 3D scene.
It is noteworthy that our method not only directly performs hierarchical clustering on 3D point-level instances to obtain consistent hierarchical semantic features, but also captures the relationships between semantic hierarchical structures and within them. This method can learn the global and local hierarchical semantic structure of 3D scenes in a comprehensive and direct manner.

During inference, similar to LangSplat~\cite{qin2024langsplat}, we follow Eq.~\ref{M} to project language embeddings from 3D to 2D, then use a scene-specific decoder $\Psi$ to recover the CLIP image embedding $\Psi(\mathbf{M}^t) \in \mathbb{R}^{D \times H \times W}$, enabling open-vocabulary queries with the CLIP text encoder.

%Additionally, we use the Hierarchical Consistency (HC) score ${s_{HC}} = \frac{{{\mathop{\rm Area}\nolimits} \left( {{\mathbf{M}^{{\rm{l - 1}}}} \cap {\mathbf{M}^{\rm{l}}}} \right)}}{{{\mathop{\rm Area}\nolimits} \left( {{\mathbf{M}^{{\rm{l - 1}}}}} \right)}}$ to further assess the part-wise semantic layering ability. 

\section{Hierarchical Datasets}
%We reconstruct two hierarchical 3D scene datasets named Hi-LERF and Hi-3DOVS to evaluate the model's capability in distinguishing layered semantics.We annotated a total of 30 images from 4 scenes in the LERF dataset~\cite{kerr2023lerf} and 50 images from 10 scenes in the 3D-OVS dataset~\cite{3DOVS}, each with three different levels of masks. The smaller masks $M^{l^w}$, $M^{l^p}$, which represent higher semantic levels, are accurately nested within the masks of the preceding layer $M^{l^p}$ and $M^{l^s}$.

%We reconstruct two hierarchical 3D scene datasets, named Hi-LERF and Hi-3DOVS, to evaluate the model's ability to differentiate hierarchical semantics. We annotate a total of $40$ images from $4$ scenes based on LERF dataset \cite{kerr2023lerf} and $100$ images from $10$ scenes based on 3D-OVS dataset \cite{3DOVS}, with each scene containing three hierarchical levels of mask sets $M^{l^w}$, $M^{l^p}$, and $M^{l^s}$. 

%We reconstructed two hierarchical 3D scene datasets, Hi-LERF and Hi-3DOVS, to evaluate the model's ability to differentiate hierarchical semantics. 
For existing 3D hierarchical semantic datasets, Blender-HS and PartNet~\cite{Mo_2019_CVPRpartnet} are limited to single objects and simple scenes. Blender-HS~\cite{ying2024omniseg3d} which is proposed by OmniSeg covers only simple plane semantic hierarchies without addressing consistent 3D features. 
%Our research shows no public dataset exists for validating hierarchical semantics and 3D consistency in complex scenes. 
Based on research, no public dataset exists to validate 3D hierarchical semantics and consistency in 3D complex scenes.
Therefore, we construct two 3D hierarchical semantics datasets, named Hi-LERF and Hi-3DOVS.
%, to evaluate models' ability to differentiate hierarchical semantics.
We annotate 40 images from 4 scenes based on the LERF~\cite{kerr2023lerf} and 60 images from 10 scenes based on the 3D-OVS~\cite{3DOVS}. Each scene contains $3$ hierarchical levels of mask sets $M^{l^w}$, $M^{l^p}$, and $M^{l^s}$, with each set containing approximately $10$ corresponding annotations.
%with each image view including nearly $10$ instance-wise and $10$ part-wise annotations.
Following the 3D semantic hierarchy tree, smaller masks represent higher semantic levels, and they are precisely nested within the previous layer's tree node masks, ensuring $M^{l^s} \subset M^{l^p} \subset M^{l^w}$. We used both automatic and manual labeling methods.
We used SAM to extract hierarchical semantic masks, analyzed their overlaps for semantic layering, and manually annotated hierarchical semantic features with labels as ground truth. 
%More details of the two hierarchical semantic datasets are in Appendix A.

\begin{figure}[!htbp]
\centering
\includegraphics[width=1\linewidth]{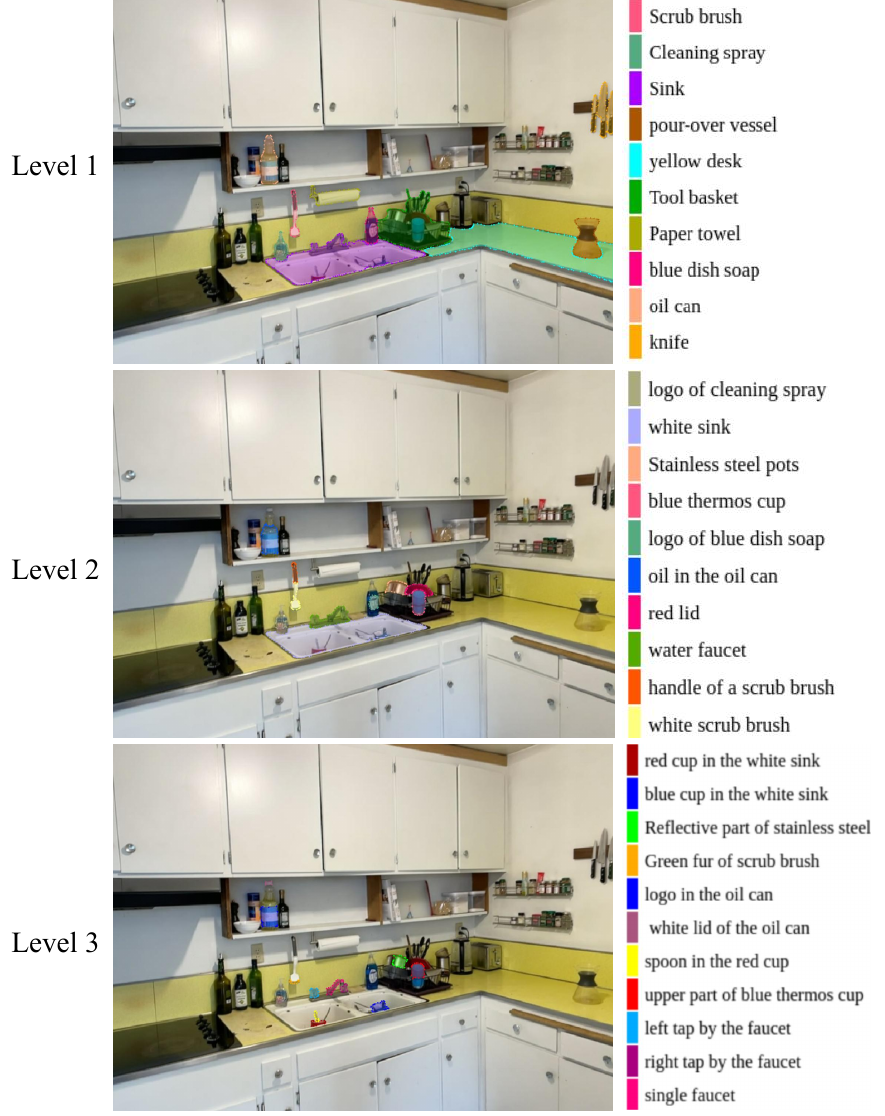}
\caption{We randomly selected several open-vocabulary queries and their corresponding labels across the three different semantic levels that we annotated.}
\label{label}
\end{figure}

Additionally, Figure~\ref{label} randomly displays a subset of our annotations and their labels across the three different semantic levels.
For the Hierarchical Consistency (HC) score, based on our 3D hierarchical semantic tree which has 3 different semantic level, we provide the detailed computation method, as below:
\begin{equation}
    \resizebox{0.9\hsize}{!}{$  {s_{H C}}=\frac{1}{(L-1) \cdot \max \left(N_l, 1\right)} \sum_{l=1}^{L-1} \sum_{i=1}^{N_l} \frac{1}{N_{i, l+1}} \sum_{j=1}^{N_{i, l+1}} \frac{\operatorname{Area}\left(M_i^l \cap M_j^{l+1}\right)}{\operatorname{Area}\left(M_i^l\right)}$}
\end{equation}
where $L$ is the total number of semantic levels. $N_l$ is the number of semantics at semantic level l that has the child nodes of the 3d hierarchical semantic tree at the next level $l+1$.
$N_{i, l+1}$ is the number of tree node semantics for $M_i^l$ at the next level $l+1$.
$M_i^l$ is a semantic at semantic level $l$.
$M_j^{l+1}$ is a mask at the next semantic level $l+1$ that contains $M_i^l$.
Area $(\cdot)$ represents the query area of the semantic.

\begin{table}[!htp]
	\centering
 \caption{We have meticulously detailed the views included in each scene of the Hi-LERF dataset.} 
      \setlength{\tabcolsep}{6pt}
  %\normalsize
	\scalebox{1}{% Please add the following required packages to your document preamble:
% \usepackage{multirow}
% \usepackage{multirow}\
\small
\begin{tabular}{llll}
\hline
\multicolumn{4}{c}{Hi-LERF}                                \\ \hline
Figurines    & Ramen        & Teatime      & Waldo kitchen \\ \hline
frame\_00016 & frame\_00006 & frame\_00002 & frame\_00010  \\
frame\_00041 & frame\_00024 & frame\_00025 & frame\_00020  \\
frame\_00060 & frame\_00042 & frame\_00043 & frame\_00033  \\
frame\_00105 & frame\_00060 & frame\_00107 & frame\_00053  \\
frame\_00122 & frame\_00065 & frame\_00116 & frame\_00066  \\
frame\_00176 & frame\_00081 & frame\_00125 & frame\_00089  \\
frame\_00152 & frame\_00094 & frame\_00129 & frame\_00125  \\
frame\_00195 & frame\_00104 & frame\_00140 & frame\_00140  \\
frame\_00226 & frame\_00119 & frame\_00158 & frame\_00154  \\
frame\_00260 & frame\_00128 & frame\_00180 & frame\_00186  \\ \hline
\end{tabular}}
% * is the implement result which contains the mean accuracy and the standard deviation by 5 runs under 5 different seeds.}  
	\label{views——lerf}
\end{table}

\begin{table}[!htp]
	\centering
 \caption{We have meticulously detailed the views included in each scene of the Hi-3DOVS dataset.} 
      \setlength{\tabcolsep}{1pt}
  %\normalsize
	\scalebox{0.65}{% Please add the following required packages to your document preamble:
% \usepackage{multirow}
% \usepackage{multirow}\
\small
\begin{tabular}{cccccccccc}
\hline
\multicolumn{10}{c}{Hi-LERF}                                                                                                  \\ \hline
bed       & bench     & blue\_sofa & covered\_desk & lawn      & office\_desk & room      & snacks    & sofa      & table     \\ \hline
frame\_01 & frame\_02 & frame\_03  & frame\_00     & frame\_01 & frame\_03    & frame\_00 & frame\_04 & frame\_02 & frame\_00 \\
frame\_06 & frame\_05 & frame\_05  & frame\_11     & frame\_03 & frame\_07    & frame\_04 & frame\_08 & frame\_04 & frame\_02 \\
frame\_11 & frame\_25 & frame\_13  & frame\_15     & frame\_09 & frame\_12    & frame\_10 & frame\_16 & frame\_10 & frame\_14 \\
frame\_20 & frame\_27 & frame\_24  & frame\_21     & frame\_13 & frame\_14    & frame\_19 & frame\_26 & frame\_15 & frame\_26 \\
frame\_28 & frame\_32 & frame\_27  & frame\_26     & frame\_29 & frame\_20    & frame\_25 & frame\_36 & frame\_22 & frame\_30 \\
frame\_36 & frame\_35 & frame\_29  & frame\_30     & frame\_35 & frame\_26    & frame\_30 & frame\_40 & frame\_27 & frame\_31 \\ \hline
\end{tabular}}
% * is the implement result which contains the mean accuracy and the standard deviation by 5 runs under 5 different seeds.}  
	\label{views——3dovs}
\end{table}

\begin{table*}[!htp]
	\centering
 \caption{Comparisons between our model and SOTA methods of semantic segmentation and localization tasks on the LERF dataset. *: Reproduced result. We color top-3 results with different colors, which are the \textbf{\colorbox[RGB]{255,179,179}{best}}, \textbf{\colorbox[RGB]{255,217,179}{second best}},
and \textbf{\colorbox[RGB]{255,255,179}{third best}}. } 
	%\resizebox{8cm}{
	\scalebox{0.9}{
\small
	\setlength{\tabcolsep}{1pt}
%\begin{tabular}{lcccccccccccc}
\begin{tabular}{lccccccccccccccc}
\hline
\multicolumn{1}{c}{\multirow{3}{*}{Method}} & \multicolumn{10}{c|}{Semantic Segmentation}                                                                                                                        & \multicolumn{5}{c}{Localization}                                                                                                                          \\ \cline{2-16} 
\multicolumn{1}{c}{}                        & \multicolumn{2}{c}{\emph{Figurines}} & \multicolumn{2}{c}{\emph{Ramen}}     & \multicolumn{2}{c}{\emph{Teatime}}   & \multicolumn{2}{c}{\emph{Waldo kitchen}} & \multicolumn{2}{c|}{Average}   & \multicolumn{1}{l}{\emph{Figurines}} & \multicolumn{1}{l}{\emph{Ramen}} & \multicolumn{1}{l}{\emph{Teatime}} & \multicolumn{1}{l}{\emph{Waldo kitchen}} & \multicolumn{1}{l}{Average} \\ \cline{2-16} 
\multicolumn{1}{c}{}                        & mIoU          & \multicolumn{1}{c|}{mBIoU}         & mIoU          & \multicolumn{1}{c|}{mBIoU}         & mIoU          & \multicolumn{1}{c|}{mBIoU}         & mIoU            & \multicolumn{1}{c|}{mBIoU}           & mIoU          & \multicolumn{1}{c|}{mBIoU}         & \multicolumn{5}{c}{Accuracy}                                                                                                                              \\ \hline
\textbf{\emph{2D-based}} &&&&&&&&&&\multicolumn{1}{c|}{}&&&&& \\
 LSeg~\cite{li2022lseg} &-&-&-&-&-&-&-&-&-&\multicolumn{1}{c|}{-}   & 8.9 & 14.1 & 33.9 & 27.3 & 21.1 \\ 
 \hline
\textbf{\emph{NeRF-based}} &&&&&&&&&&\multicolumn{1}{c|}{}&&&&& \\
LERF~\cite{kerr2023lerf}                                        & 33.5          & 30.6          &28.3          &14.7          & 49.7          & 42.6          & 37.9            & 28.4            & 37.2          & \multicolumn{1}{c|}{29.3}          & 75.0                          & 62.0                        & 84.8                        & 72.7                              & 73.6                        \\
3D-OVS~\cite{3DOVS}     &44.8 & -&28.7 & -& 56.1& -& 39.3&- &- &\multicolumn{1}{c|}{-}   & 77.3& 70.2&87.7 & 45.6& -\\
Laser~\cite{miao2025laser}     & 63.5&- & 44.6& -& 62.4&- & \cellcolor{top2}41.3& -& -& \multicolumn{1}{c|}{-}  & 76.9& 76.1 &88.7 &79.1 &- \\
OmniSeg3D-NeRF*~\cite{ying2024omniseg3d}                                   & 30.9          & 26.4          & 25.3          & 13.6          & 42.6         & 40.7          & 33.5            & 24.2            & 31.9          & \multicolumn{1}{c|}{23.7}         & 66.9                          & 55.8                      &72.9                       & 66.3                              & 64.8                       \\ 
Open3DRF-NeRF~\cite{wu2024opengaussian}                                & -        & -             & -          & -             & -         & -             & -         &           & 46.4      & \multicolumn{1}{c|}{45.4 }             & -                             & -                         & -                           & -                                 & -                           \\ 
\hline
\textbf{\emph{3DGS-based}} &&&&&&&&&&\multicolumn{1}{c|}{}&&&&& \\
LEGaussians~\cite{shi2024LEGaussians}                                & 18.0         & -             &15.8           & -             &19.3       & -             & 11.8         & -               & 16.2        &\multicolumn{1}{c|}{-}            & -                             & -                         & -                           & -                                 & -                           \\ 
OmniSeg3D-GS*~\cite{ying2024omniseg3d}                                   & 31.4          & 26.8          & 25.9          & 14.3      & 43.7         & 41.2         & 34.4           & 24.7           &33.9         & \multicolumn{1}{c|}{26.8}         & 67.3                          & 56.4                    &73.8                       & 66.9                             & 66.1                    \\
OpenGaussian~\cite{wu2024opengaussian}                                & 39.3          & -             & 31.0            & -             & 60.4          & -             & 22.7            & -               & 38.4         & \multicolumn{1}{c|}{-}             & -                             & -                         & -                           & -                                 & -                           \\ 
SLAG~\cite{szilagyi2025slag}                                   & 48.1     & -             & 24.8        & -             &56.2       & -             &27.6        & -               & 39.1     & \multicolumn{1}{c|}{-}             & -                             & -                         & -                           & -                                 & -                           \\ 
Open3DRF-GS~\cite{lee2025rethinkingopenvocabularysegmentationradiance}                                & -        & -             & -          & -             & -         & -             & -         &           & 44.4  & \multicolumn{1}{c|}{44.6}             & -                             & -                         & -                           & -                                 & -                           \\ 
SuperGSeg~\cite{liang2024supergseg}   & 43.7 & 60.7 &  18.1&  23.9& 55.3 &78.0  & 26.7 &\cellcolor{top2}45.5  & 35.5 &\multicolumn{1}{c|}{\cellcolor{top2}52.0} & - & - &-  & - &-   \\
Semantic Gaussians~\cite{guo2024semantic}   &  -& - &-  &-  & - & - & - &-  & - &\multicolumn{1}{c|}{-}  & 83.1 &76.8  & 89.8 &90.0  &  85.2 \\
FastLGS~\cite{ji2025fastlgs}  &  -& - &-  &-  & - & - & - &-  & - &\multicolumn{1}{c|}{-}  & \cellcolor{top2} 91.4&\cellcolor{top2} 84.2 &\cellcolor{top2} 95.0 &\cellcolor{top2} 96.2 &\cellcolor{top2} 91.7 \\ 
FreeGS~\cite{zhang2025bootstraping}  & 62.6 & - & 77.5 &-  & 68.5 & - & - & - &\cellcolor{top3}52.2 & \multicolumn{1}{c|}{-} & - & - &-  & - & -  \\
LangSplat~\cite{qin2024langsplat}                                   & \cellcolor{top3}44.7          & \cellcolor{top3}41.9          & \cellcolor{top3}51.2          & \cellcolor{top3}48.8          & \cellcolor{top3}65.1          & \cellcolor{top3}60.8          & \cellcolor{top3}44.5            & \cellcolor{top3}39.1            & 51.5          & \multicolumn{1}{c|}{47.8}          & 80.4                          & 73.2                      & 88.1                        & 95.5                              &84.3                        \\ 
Gaussian Grouping~\cite{ye2024gaussiangroupingsegmentedit}                                   & \cellcolor{top2}69.7          & \cellcolor{top2}67.9         & \cellcolor{top2}77.0       & \cellcolor{top2}68.7         & \cellcolor{top2}71.7         & \cellcolor{top2}66.1         & -           & -            &\cellcolor{top2} 54.6        & \multicolumn{1}{c|}{\cellcolor{top3}50.7}          & \cellcolor{top3}84.7                         & \cellcolor{top3}80.2                      &\cellcolor{top3} 91.3                      & \cellcolor{top3}96.1                             & \cellcolor{top3}88.1                    \\
\hline
\textbf{Ours}                               & \textbf{\cellcolor{top1}71.5} & \textbf{\cellcolor{top1}69.3} & \textbf{\cellcolor{top1}78.8} & \multicolumn{1}{c}{\textbf{\cellcolor{top1}72.6}} & \textbf{\cellcolor{top1}73.7} & \textbf{\cellcolor{top1}67.3} & \textbf{\cellcolor{top1}58.8}   & \textbf{\cellcolor{top1}51.3}   & \textbf{\cellcolor{top1}68.4} &  \multicolumn{1}{c|}{\textbf{\cellcolor{top1}65.1}} & \textbf{\cellcolor{top1}93.2}                 & \textbf{\cellcolor{top1}85.7}             & \textbf{\cellcolor{top1}95.2}               & \textbf{\cellcolor{top1}96.8}                     & \textbf{\cellcolor{top1}92.7}               \\ \hline
\end{tabular}}
	\label{lerf}
\end{table*}

\begin{table*}
	\centering
 \caption{Comparisons between our model and SOTA methods of semantic segmentation task on the 3D-OVS datasets.} 
	%\resizebox{8cm}{
	\scalebox{0.9}{% Please add the following required packages to your document preamble:
% \usepackage{multirow}
% \usepackage{multirow}\
\small
	\setlength{\tabcolsep}{23pt}
%\begin{tabular}{lcccccccccccc}
\begin{tabular}{lcccccc}
\hline
\multicolumn{1}{c}{\multirow{2}{*}{Method}} & \multicolumn{6}{c}{Semantic Segmentation(mIoU)} \\ \cline{2-7} 
\multicolumn{1}{c}{}                        & \emph{Bed}   & \emph{Bench}  & \emph{Room}  & \emph{Sofa}  & \emph{Lawn} & Overall \\ \hline
\textbf{\emph{2D-based}} \\
LSeg~\cite{li2022lseg}                                        & 56.0 & 6.0   & 19.2 & 4.5  & 17.5 & 20.6    \\
ODISE~\cite{Xu_2023_CVPRODISE}                                       & 52.6 & 24.1  & 52.5 & 48.3 & 39.8 & 43.5    \\
OV-Seg~\cite{Liang_2023_CVPRseg}                                      & 79.8 & 88.9  & 71.4 & 66.1 & 81.2 & 77.5    \\ \hline
\textbf{\emph{NeRF-based}} \\
FFD~\cite{kobayashi2022decomposingFFD}                                & 56.6 & 6.1   & 25.1 & 3.7  & 42.9 & 26.9    \\
LERF~\cite{kerr2023lerf}                                        & 73.5 & 53.2  & 46.6 & 27.0 & 73.7 & 54.8    \\
Open3DRF-NeRF &-&-&-&-&-&77.5 \\
3D-OVS~\cite{3DOVS}                                      & 89.5& 89.3 &92.8& 74.0 & 88.2 & 86.8  \\
OmniSeg3D-NeRF*~\cite{ying2024omniseg3d}  & 89.7& 90.2& 92.0& 75.3&88.5 &87.1 \\
Laser~\cite{miao2025laser}                                        & 91.4 & 88.3  &  85.9 &  86.0  & 88.5 & 88.1   \\
\hline
\textbf{\emph{3DGS-based}} \\
LEGaussians~\cite{shi2024LEGaussians}  & 56.8& 28.9&57.2&52.6& 44.1&47.9 \\
OmniSeg3D-GS*~\cite{ying2024omniseg3d}  & 89.9 &91.0 &92.3 &76.1 &89.6 & 87.8\\
Open3DRF-GS &-&-&-&-&-&77.5 \\
FMGS~\cite{zuo2025fmgs} &80.6&84.5&87.9&\cellcolor{top2}90.8&92.6&87.3 \\
Gaussian Grouping~\cite{Kim_2024_CVPRgrop}                           & \textbf{\cellcolor{top1}97.3}& 73.7  & 79.0 & 68.1 & \cellcolor{top2}96.5 & 82.9    \\
CGC~\cite{silva2024contrastivegaussianclusteringweakly}                                         & \cellcolor{top3}95.2 & \cellcolor{top2}96.1  & 86.8 & 67.5 & 91.8 &87.5    \\
FastLGS~\cite{ji2025fastlgs} &94.7 & \cellcolor{top3}95.1 &\cellcolor{top2}95.3 & \cellcolor{top3}90.6 &93.9 &\cellcolor{top2}95.1  \\
LangSplat~\cite{qin2024langsplat}                                   & 92.5 & 94.2  & \cellcolor{top3}94.1 & 90.0 & \cellcolor{top3}96.1 & \cellcolor{top3}93.4    \\
\hline
\textbf{Ours}                                        & {\cellcolor{top2}95.9} & \textbf{\cellcolor{top1}97.3}  & \textbf{\cellcolor{top1}96.7} & \textbf{\cellcolor{top1}92.8} & \textbf{\cellcolor{top1}97.6} & \textbf{\cellcolor{top1}96.1}    \\ \hline
\end{tabular}}
	\label{3d}
\end{table*}

\section{Experiment}
\subsection{Datasets and Implementation Details}

%The specific calculation method is as follows: ${s_{HC}} = \frac{{{\mathop{\rm Area}\nolimits} \left( {{M^{{\rm{l - 1}}}} \cap {M^{\rm{l}}}} \right)}}{{{\mathop{\rm Area}\nolimits} \left( {{M^{{\rm{l - 1}}}}} \right)}}$.

\subsubsection{Datasets}
We evaluated our model on 6 public datasets and 2 constructed hierarchical datasets for semantic and instance segmentation, and localization tasks. %The LERF dataset~\cite{kerr2023lerf}, designed for 3D object localization, consists of 4 complex outdoor 3D scenes.We used the LERF-Mask from LangSplat~\cite{qin2024langsplat} to evaluate semantic segmentation.

\textbf{LERF} dataset~\cite{kerr2023lerf}, designed for 3D object localization, consists of complex outdoor 3D scenes captured using the Polycam app on an iPhone, covering 4 distinct scenes.
To adapt LERF for evaluating semantic segmentation capabilities, we employed the LERF-Mask from LangSplat~\cite{qin2024langsplat}, annotating masks with more challenging text queries to improve segmentation and localization assessment.
 
\begin{table*}[!htp]
	\centering
    \small
 \caption{Experiments on Replica~\cite{straub2019replica} for semantic segmentation. *: Reproduced result.  } 
      \setlength{\tabcolsep}{0.1pt}
	\scalebox{0.95}{% Please add the following required packages to your document preamble:
% \usepackage{multirow}
% \usepackage{multirow}\、
\small
\begin{tabular}{lcccccccccccccccc}
\hline
\multicolumn{1}{c}{\multirow{3}{*}{Method}} & \multicolumn{16}{c}{Replica}                                                                                                                                                                                                            \\ \cline{2-17} 
\multicolumn{1}{c}{}                        & \multicolumn{2}{c}{office0} & \multicolumn{2}{c}{office1} & \multicolumn{2}{c}{office2} & \multicolumn{2}{c}{office3} & \multicolumn{2}{c}{office4} & \multicolumn{2}{c}{room0} & \multicolumn{2}{c}{room1} & \multicolumn{2}{c}{room2} \\ \cline{2-17} 
\multicolumn{1}{c}{}                        & mIoU↑        & mAcc↑        & mIoU↑        & mAcc↑        & mIoU↑        & mAcc↑        & mIoU↑        & mAcc↑        & mIoU↑        & mAcc↑        & mIoU↑       & mAcc↑       & mIoU↑       & mAcc↑       & mIoU↑       & mAcc↑       \\ \hline
\textbf{\emph{2D-based}} \\
LSeg~\cite{li2022lseg} & 1.05 & 6.73 &0.92 &4.78 & 5.31& 9.72& 3.62& 11.92& 1.93& 4.91& 4.92&14.38 &4.33 & 13.94&1.78 & 15.41\\ \hline

\textbf{\emph{NeRF-based}} \\
LERF~\cite{kerr2023lerf}                                        & 11.56        & 35.82        & 12.95        & 37.81        & 14.92        & 39.41        & 12.60        & 37.20        & 8.20         & 16.30        & 12.80       & 29.70       & 12.80       & 29.70       & 40.13       & 52.42       \\
3D-OVS~\cite{3DOVS}                                      & 12.83        & 38.21        & 13.06        & 38.66        & 15.84        & 38.63        & 12.10        & 36.10        & 15.80        & 28.90        & 12.50       & 40.10       & 13.10       & 30.20       & 41.04       & 52.97       \\
OmniSeg3D-NeRF*~\cite{ying2024omniseg3d} &14.31 &38.02 &15.73 &42.88 &17.89 &21.84 &15.76 &40.93 & 15.41&55.32 &21.59 &38.76 &17.83 & 30.93&40.55 & 52.62\\
Laser~\cite{miao2025laser}                                       & 16.82        & {\cellcolor{top2}40.25}        & {\cellcolor{top3}19.47}        & {\cellcolor{top2}48.62}        & 23.58        & 52.74        & \cellcolor{top3}18.70        & {\cellcolor{top2}64.30}        & \cellcolor{top3}39.70        & \cellcolor{top3}62.50        & \cellcolor{top3}24.30       & {\cellcolor{top2}54.70}       & \cellcolor{top3}25.00       & {\cellcolor{top2}45.90}       & {\cellcolor{top2}45.53}       & {\cellcolor{top2}56.31}       \\ \hline
\textbf{\emph{3DGS-based}} \\
LangSplat~\cite{qin2024langsplat}                                       & 2.43         & 11.09        & 2.10         & 1.36         & 5.68         & 10.70        & 4.65         & 13.99        & 1.49         & 2.37         & 3.86        & 12.82       & 4.08        & 12.24       & 0.92        & 10.05       \\
OmniSeg3D-GS*~\cite{ying2024omniseg3d} & 15.42& 38.37& 17.53& \cellcolor{top3}44.32&18.42 & 23.52&16.41 &41.82 & 17.97& 17.32&23.01 & 40.74&19.84 &30.97 & 41.03&53.93 \\
OpenGaussian~\cite{wu2024opengaussian}                                      & {\cellcolor{top3}17.20}        & 36.54        & {\cellcolor{top2}23.13}        & 35.11        & {\cellcolor{top2}43.72}        & \cellcolor{top3}66.38        & {\cellcolor{top2}42.36}       & \cellcolor{top3}42.64        & {\cellcolor{top2}61.33}        & {\cellcolor{top2}69.62}        & {\cellcolor{top2}31.45}       & \cellcolor{top3}41.74       & {\cellcolor{top2}40.36}       & \cellcolor{top3}31.72       & \cellcolor{top3}42.14       & \cellcolor{top3}54.10       \\
Gaussian Grouping~\cite{ye2024gaussiangroupingsegmentedit}                                        & {\cellcolor{top2}19.58}        & {\cellcolor{top3}38.42}        & -         & -    & \cellcolor{top3}32.77        & {\cellcolor{top2}74.48}        & 10.18        & 26.17        & 30.29        & 45.67        & 13.08       & 36.21       & 17.81       & 31.57       & 17.06       & 24.17       \\ \hline
\textbf{Ours}                                        & \textbf{\cellcolor{top1}25.91}        & \textbf{\cellcolor{top1}51.93}        & \textbf{\cellcolor{top1}24.85}        & \textbf{\cellcolor{top1}49.91}        & \textbf{\cellcolor{top1}45.83}        & \textbf{\cellcolor{top1}75.81}        & \textbf{\cellcolor{top1}43.09}        & \textbf{\cellcolor{top1}47.63}        & \textbf{\cellcolor{top1}65.01}        & \textbf{\cellcolor{top1}82.94}        & \textbf{\cellcolor{top1}41.77}       & \textbf{\cellcolor{top1}61.04}       & \textbf{\cellcolor{top1}65.63}       & \textbf{\cellcolor{top1}84.89}       & \textbf{\cellcolor{top1}48.53}       & \textbf{\cellcolor{top1}65.01}       \\ \hline
\end{tabular}
}
%\setlength{\abovecaptionskip}{-0.00cm}
%\captionsetup{font={small}}
	\label{replica1}
\end{table*}

% we followed the LERF-Mask from LangSplat \cite{qin2024langsplat}, conducting a more robust evaluation based on masks annotated with more complex textual queries.
%We evaluate our model on two datasets across scene segmentation and localization tasks. The LERF dataset~\cite{kerr2023lerf}, designed for 3D object localization tasks, comprises complex outdoor 3D scenes collected via the Polycam app on an iPhone, encompassing 10 different scenes. 
%The 3DOVS dataset~\cite{3DOVS} comprises ten scene collections featuring long-tail objects captured in varying poses and backgrounds. This dataset is specifically designed for open-vocabulary 3D semantic segmentation and includes a comprehensive list of categories.
%To extend this dataset for evaluating segmentation capabilities, we followed LangSpats~\cite{qin2024langsplat} by annotating masks for text queries, allowing for the assessment of open-vocabulary 3D semantic segmentation. Following LERF~\cite{kerr2023lerf}, we report localization accuracy for 3D object localization tasks and IoU results for 3D semantic segmentation tasks.
%To adapt this dataset to LERF-Mask for evaluating semantic segmentation capabilities, we followed LangSplat~\cite{qin2024langsplat} by annotating masks based on more complex text queries for better evaluation.
%enabling the assessment of open-vocabulary 3D semantic segmentation. 

\begin{table*}[!htp]
	\centering
    \small
 \caption{Experiments on ScanNetv2~\cite{dai2017scannet} for semantic segmentation, and on ScanNet200~\cite{rozenberszki2022languagescannet200} for semantic segmentation and instance segmentation. *: Reproduced result. } 
      \setlength{\tabcolsep}{6pt}
	\scalebox{0.9}{% Please add the following required packages to your document preamble:
% \usepackage{multirow}
% \usepackage{multirow}\、
\small
\begin{tabular}{l|cccccc|ccccc}
\hline
\multirow{3}{*}{Method} & \multicolumn{6}{c|}{ScanNet}                                                                      & \multicolumn{5}{c}{ScanNet200}                                                        \\ \cline{2-12} 
                        & \multicolumn{2}{c}{19 classes} & \multicolumn{2}{c}{15 classes} & \multicolumn{2}{c|}{10 classes} & \multicolumn{2}{c}{semantic segmentation} & \multicolumn{3}{c}{instance segmentation} \\ \cline{2-12} 
                        & mIoU↑          & mAcc↑         & mIoU↑          & mAcc↑         & mIoU↑          & mAcc↑          & mIoU↑               & mAcc↑               & AP      & { AP@25}    & { AP@50}    \\ \hline
                        \textbf{\emph{2D-based}} &&&&&&\multicolumn{1}{c|}{}&&&&&\\
                    Lseg~\cite{li2022lseg} &0.1 & -&0.4 & -&0.9 &- & 1.6&3.3& -&- &- \\ 
        OpenScene~\cite{Peng_2023_CVPRopenscene} &\cellcolor{top2}43.6 & -&\cellcolor{top2}51.3 & -&\cellcolor{top2}58.3&- &  \cellcolor{top3}6.4&\cellcolor{top3}12.2& 8.5&- &- \\ 
                    \hline
                    \textbf{\emph{NeRF-based}} &&&&&&\multicolumn{1}{c|}{}&&&&& \\
                
                    LERF~\cite{kerr2023lerf} &15.8 &25.3 & 21.5& 35.5&36.5 &48.1 &5.8 & 10.6& 13.2& 17.5& 26.3\\
                                        3D-OVS~\cite{3DOVS} & 17.3&29.3 &24.8 &38.3 &38.4 & 54.7&6.3 & 11.2&14.9 & 19.7&28.4 \\
                    OmniSeg3D-NeRF*~\cite{ying2024omniseg3d} &21.1 &38.4 & 27.9& 41.2& 40.3&62.4 &7.5 &11.5 &17.6 & 21.5& 31.6\\\hline
                    \textbf{\emph{3DGS-based}} &&&&&&\multicolumn{1}{c|}{}&&&&&\\ 
LangSplat~\cite{qin2024langsplat}               & 2.0            & 9.2           & 4.9            & 14.6          & 8.0            & 23.9           & 2.5                 & 6.4                 & \cellcolor{top3}19.5    & 21.3           & 28.6           \\
LEGaussians~\cite{shi2024LEGaussians}          & 1.6            & 7.9           & 4.6            & 16.1          & 7.7            & 24.9           & -                   & -                   & -       & -              & -              \\
OmniSeg3D-GS*~\cite{ying2024omniseg3d} & 23.5&41.3 &32.6 & 43.7& 42.9&63.8 & 8.2& 13.3& 11.4&\cellcolor{top3}23.0 &\cellcolor{top3}32.7 \\
OpenGaussian~\cite{wu2024opengaussian}            & 30.1           & 46.5          & 38.1           & \cellcolor{top3}56.8          & 49.7           & \cellcolor{top3}71.4           & \cellcolor{top2}10.5                & \cellcolor{top2}15.1                & \cellcolor{top2}20.8    & \cellcolor{top2}25.7           & \cellcolor{top2}37.4           \\
Dr. Splat~\cite{jun2025dr}              & 29.6           & \cellcolor{top3}47.7          & \cellcolor{top3}38.2           & \cellcolor{top2}60.4          & \cellcolor{top3}50.2           & \cellcolor{top2}73.5           & -                   & -                   & -       & -              & -              \\ 
SLAG~\cite{szilagyi2025slag}                     &\cellcolor{top3} 31.3          & \cellcolor{top2}49.8          &30.7          & 50.0         &48.3    & \cellcolor{top2}73.5          & -                   & -                   & -       & -              & -              \\\hline
\textbf{Ours}                    & \textbf{\cellcolor{top1}47.9}           & \textbf{\cellcolor{top1}61.3}          & \textbf{\cellcolor{top1}53.7}           & \textbf{\cellcolor{top1}67.4}          & \textbf{\cellcolor{top1}60.1}          & \textbf{\cellcolor{top1}74.9}           & \textbf{\cellcolor{top1}18.7}                & \textbf{\cellcolor{top1}24.5}               & \textbf{\cellcolor{top1}24.5}    & \textbf{\cellcolor{top1}31.1}           & \textbf{\cellcolor{top1}43.9}           \\ \hline
\end{tabular}
}
%\setlength{\abovecaptionskip}{-0.00cm}
%\captionsetup{font={small}}
	\label{scannet3}
\end{table*}

\begin{table}[!htp]
	\centering
    \small
 \caption{Experiments Waymo~\cite{sun2020scalabilitywaymo} for semantic segmentation.} 
      \setlength{\tabcolsep}{20pt}
	\scalebox{1}{% Please add the following required packages to your document preamble:
% \usepackage{multirow}
% \usepackage{multirow}\、
\small
\begin{tabular}{lc}
\hline
\multicolumn{1}{c}{\multirow{2}{*}{Method}} & Waymo \\ \cline{2-2} 
\multicolumn{1}{c}{}                        & mIoU↑ \\ \hline
\textbf{\emph{2D-based}} \\
LSeg~\cite{li2022lseg}                                        & 15.7  \\ \hline
\textbf{\emph{NeRF-based}} \\
LERF~\cite{kerr2023lerf}                                        & 46.8  \\
3D-OVS~\cite{3DOVS}                                       & 53.1  \\
OmniSeg3D-NeRF*~\cite{ying2024omniseg3d} & 54.7 \\
Laser~\cite{miao2025laser}                                        & 58.3  \\ 
\hline
\textbf{\emph{3DGS-based}} \\
LangSplat~\cite{qin2024langsplat}                                    & 64.2  \\
OmniSeg3D-GS*~\cite{ying2024omniseg3d} & 65.1 \\
OpenGaussian~\cite{wu2024opengaussian}                                   & \cellcolor{top3}67.3  \\

Gaussian Grouping~\cite{ye2024gaussiangroupingsegmentedit}                                     & \cellcolor{top2}68.8  \\ \hline
\textbf{Ours}                                        & \textbf{\cellcolor{top1}69.5}  \\ \hline
\end{tabular}
}
%\setlength{\abovecaptionskip}{-0.00cm}
%\captionsetup{font={small}}
	\label{waymo1}
\end{table}
\textbf{3DOVS} dataset~\cite{3DOVS}, designed for open-vocabulary 3D semantic segmentation, contains 10 scenes with various long-tail object classes.
Following OpenGaussian, we randomly selected 10 scenes from ScanNet~\cite{dai2017scannet} for evaluation: scene0000, scene0062, scene0070, scene0097, scene0140, scene0200, scene0347, scene0400, scene0590, and scene0645.
For text queries, we used 19 ScanNet-defined categories: wall, floor, cabinet, bed, chair, sofa, table, door, window, bookshelf, picture, counter, desk, curtain, refrigerator, shower curtain, toilet, sink, and bathtub. A subset of 15 categories excludes picture, refrigerator, shower curtain, and bathtub, while a further reduced set of 10 omits cabinet, counter, desk, curtain, and sink.

%Metrics: We evaluate mIoU metric for semantic segmentation task.

\textbf{ScanNetv2} dataset~\cite{dai2017scannet} provides images, point clouds, and 3D point-level semantic labels. Like OpenGaussian~\cite{wu2024opengaussian}, we use 19, 15, and 10 categories for text queries.

\textbf{ScanNet200}~\cite{rozenberszki2022languagescannet200} consists of 1,201 training and 312 validation scenes spanning 198 object categories, making it ideal for assessing real-world open-vocabulary scenarios with a long-tailed distribution.

%etrics: We evaluate mIoU metric for semantic segmentation task.

%Metrics: For the ScanNet dataset, we conducted evaluations under different settings, including 19-class, 15-class, and 10-class configurations, and reported mIoU and mAcc for the semantic segmentation task.Additionally, for the ScanNet200 dataset, we performed both semantic segmentation and instance segmentation. We report mIoU and mAcc for semantic segmentation, and AP, AP@25, and AP@50 for instance segmentation.

%ScanNet200~\cite{rozenberszki2022languagescannet200} includes 312 indoor scans in its validation set and  1,201 scans in training, covering 200 object categories. 
%The Waymo~\cite{sun2020scalabilitywaymo} for 3D segmentation includes 23,691 training, 5,976 validation, and 2,982 test samples. 

\textbf{Waymo}~\cite{sun2020scalabilitywaymo} dataset for 3D semantic segmentation comprises 23,691 training samples, 5,976 validation samples, and 2,982 testing samples [37]. Each sample includes a 64-beam point cloud and RGB images from five cameras: front, front-left, front-right, side-left, and side-right. Since the Waymo ego-vehicle lacks a rear camera, points outside the field of view introduce additional challenges for multimodal segmentation.

%Metrics: We evaluate mIoU for semantic segmentation task.
%Replica~\cite{straub2019replica} has 8 evaluation scenes for instance segmentation.

\begin{figure*}[!htp]
\centering
\includegraphics[width=1\linewidth]{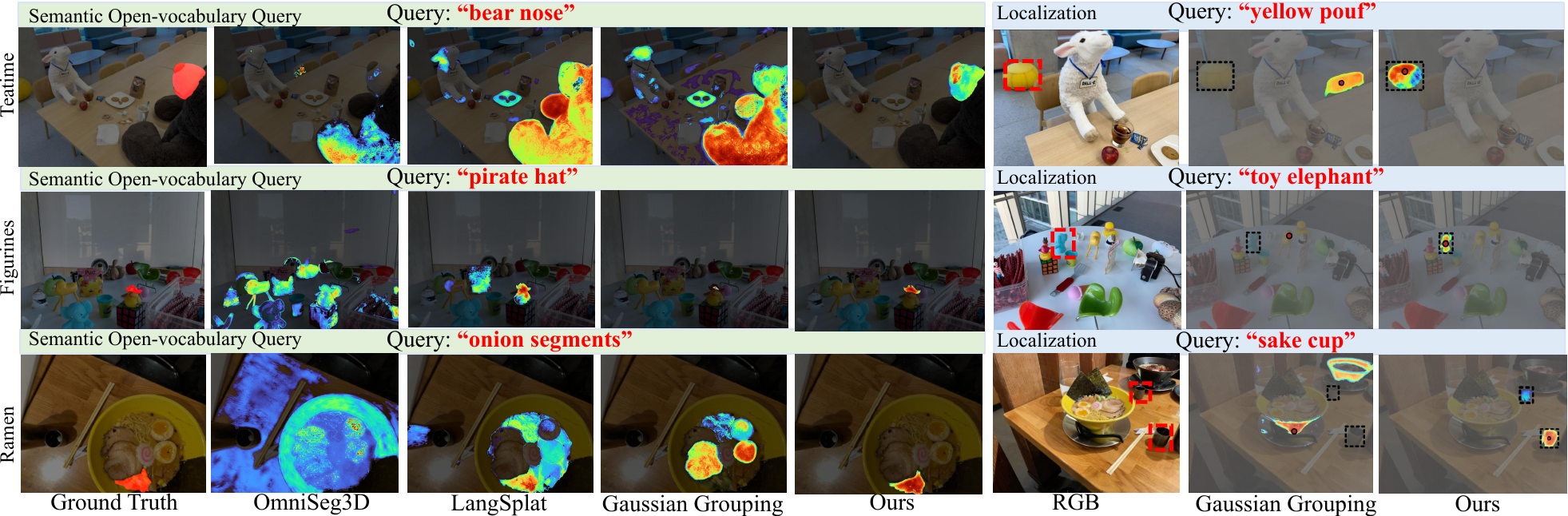}
\caption{Comparison of open-vocabulary semantic query (left) and semantic localization task (right) on the LERF dataset.}
\label{viz_seg}
\end{figure*}
\begin{figure*}[!htp]
\centering
\includegraphics[width=0.93\linewidth]{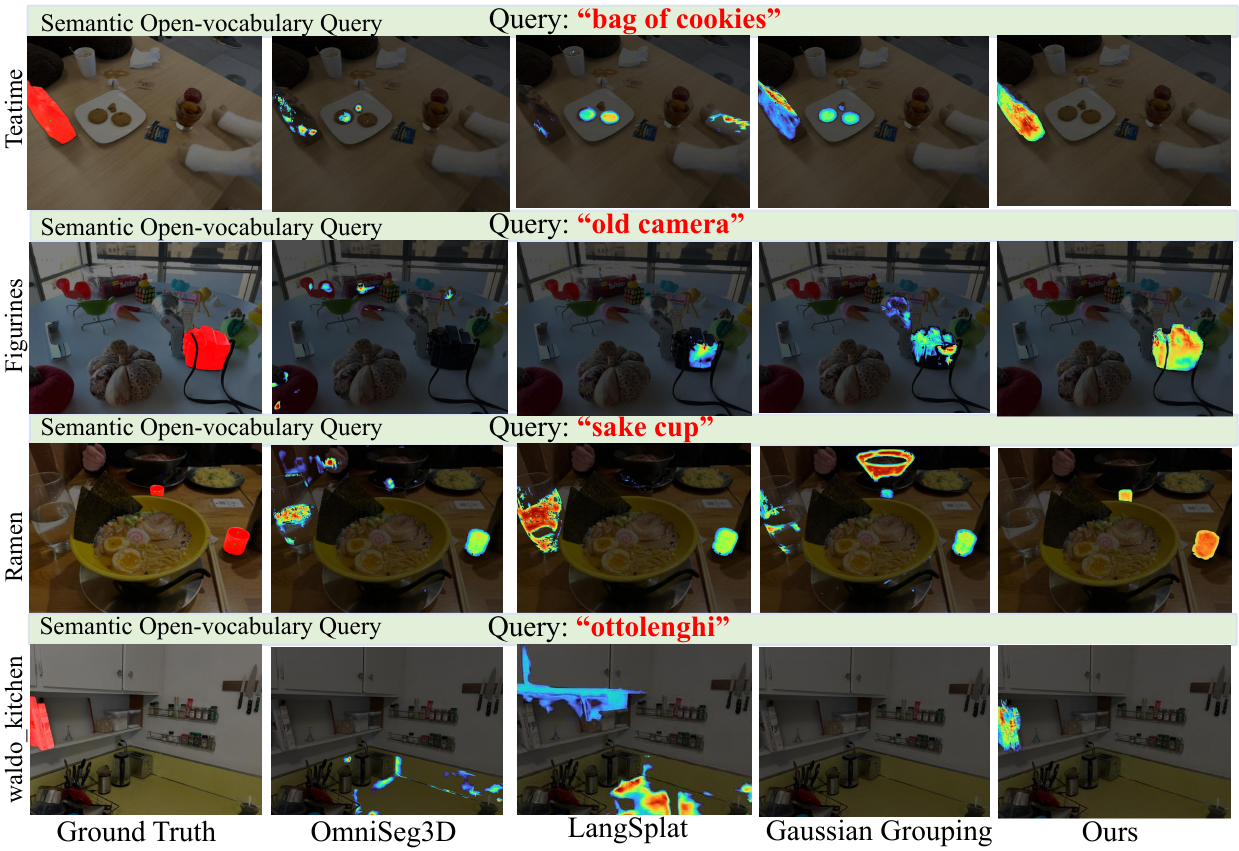}
\caption{More comparison of open-vocabulary semantic query on the LERF dataset.}
\label{viz_seg2}
\end{figure*}
\textbf{Replica}~\cite{straub2019replica} is a synthetic dataset derived from high-fidelity real-world data, featuring ground-truth 3D meshes with semantic annotations. It includes 8 evaluation scenes and 48 object classes.
%We report localization accuracy for 3D semantic localization task and IoU scores for the semantic segmentation task.
Unlike others that generate prediction masks for all classes, we focus solely on the queried semantic masks.

%Metrics: We report mIoU and mAcc scores for the semantic segmentation task. 
%Following LangSplat~\cite{qin2024langsplat}, we generated corresponding masks based on querying and evaluated performance using the mIoU metric.
%Implementation settings and introduction of datasets are in Appendix B and F.

For a fair comparison, we used all test scenes from the LERF-OVS annotated by LangSplat~\cite{qin2024langsplat} and 3D-OVS~\cite{3DOVS} datasets without modifications. For the custom two datasets, we annotated all the scenes in datasets with hierarchical semantics for testing and comparisons.%, detailed introductions are in {Appendix A}.
\subsubsection{Metrics}
We adopt a variety of evaluation metrics to comprehensively compare the performance of our model against other approaches.
For the LERF dataset~\cite{kerr2023lerf}, we compute mIoU and mBIoU for the semantic segmentation task, and accuracy for the object localization task.
For the 3DOVS dataset~\cite{3DOVS}, we evaluate the mIoU metric for semantic segmentation.
For the ScanNet dataset~\cite{dai2017scannet}, we conduct evaluations under different settings, including 19-class, 15-class, and 10-class configurations, and report mIoU and mAcc for the semantic segmentation task.
For the ScanNet200 dataset~\cite{dai2017scannet}, we perform both semantic segmentation and instance segmentation, reporting mIoU and mAcc for the former, and AP, AP@25, and AP@50 for the latter.
For the Waymo dataset~\cite{sun2020scalabilitywaymo}, we evaluate mIoU for semantic segmentation.
For the Replica dataset~\cite{straub2019replica}, we report both mIoU and mAcc scores for the semantic segmentation task.

\begin{figure*}[!htp]
\centering
\includegraphics[width=1\linewidth]{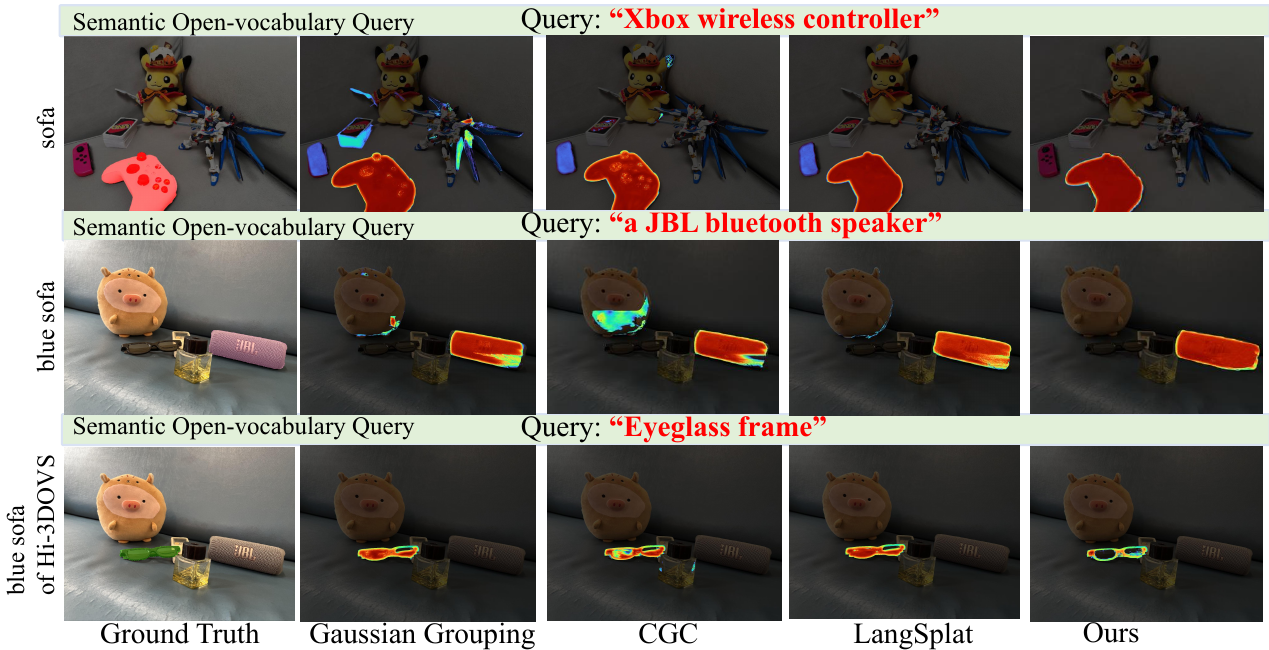}
\caption{More comparison of open-vocabulary semantic query on the 3D-OVS and Hi-3DOVS datasets.}
\label{viz_3dovs}
\end{figure*}
\begin{figure*}[!htp]
\centering
\includegraphics[width=0.95\linewidth]{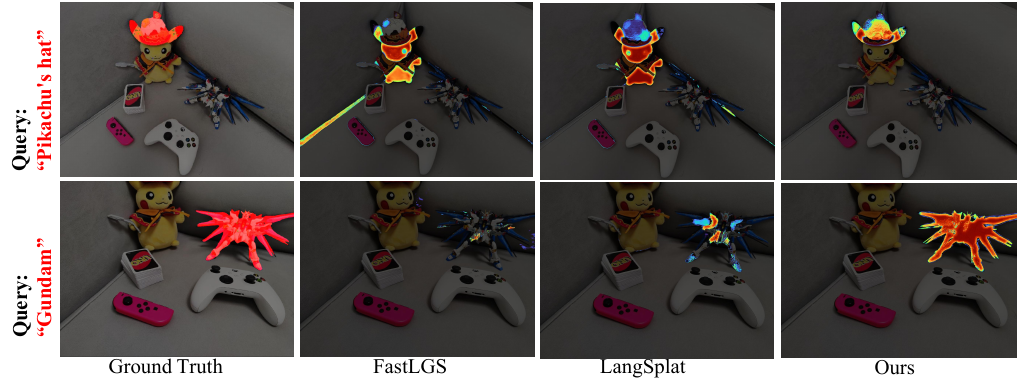}
    \caption{Comparison of open-vocabulary semantic query on the ``sofa" scene of 3D-OVS and Hi-3DOVS datasets.}
\label{viz_loca}
\end{figure*}
\subsubsection{Implementation Details}
We conduct experiments on a single NVIDIA 3090 GPU using PyTorch. Consistent with the official 3D-GS~\cite{kerbl20233gs} setup, we utilize original RGB scenes and maintain original parameter settings. We freeze the remaining parameters of 3D-GS and only train the Gaussian semantic features for 30,000 iterations.
We employ the ViT-H from SAM~\cite{Kirillov_2023_ICCVsam} to extract whole, part, and sub-part masks. For image-language features and the auto-encoder, we use the OpenCLIP ViT-B/16 model~\cite{schuhmann2021laion400mopendatasetclipfiltered} and an MLP, respectively. We first obtain 512-D features via CLIP, which are then compressed into 3-D latent features using the auto-encoder. The loss term parameters $\lambda_1$ and $\lambda_2$ are set to $e$-$6$ and $e$-$5$, respectively. The $\theta$ and $\Omega$ are set $0.9$ and $10$. %More settings and ablations are in Appendix C. 
Following Langsplat~\cite{qin2024langsplat}, for each text query, we utilize the trained 3D language Gaussians to generate relevancy maps. Various strategies are then employed to select the optimal semantic level and obtain predictions for different tasks.
For the open-vocabulary query localization task on the LERF dataset, to mitigate the impact of outliers, we initially apply a mean convolution filter with a size of 20 to smooth the relevancy map values. We then select the map with the highest smoothed relevancy score and use the corresponding position as the final prediction.
For the open-vocabulary query semantic segmentation task on the LERF dataset, a similar approach is taken. We apply a mean filter of size 20 to smooth the relevancy maps and then proceed with binary mask prediction. The relevancy scores are first normalized, and a threshold is used to obtain a binary image as the final prediction mask. The same method is applied to the Hi-LERF dataset.
For the open-vocabulary query semantic segmentation task on the LERF and Hi-LERF datasets, each class query yields a relevancy map. We apply a relevancy threshold of 0.4, setting scores below 0.4 to 0 and scores above 0.4 to 1, thereby producing a binary map. The average relevancy score within the mask region of each map is computed, and this score is used to determine the final predicted binary map.

\subsubsection{Open-vocabulary Query}
For open-vocabulary query benchmarking, we follow  OmniSeg3D~\cite{ying2024omniseg3d} and LangSplat~\cite{qin2024langsplat}, where the model receives a 2D query point 
$q$ from a given frame 
$I$ as input and outputs a dense 2D score map. Semantic query masks are obtained by setting corresponding thresholds. 
We utilized two evaluation metrics: mIoU scores across three hierarchical levels and Hierarchical Consistency scores. To assess the model's instance-wise semantic hierarchy capability, we calculated IoU accuracy at three distinct semantic levels: $l_1$, $l_2$, and $l_3$, and their average.
Additionally, we use the Hierarchical Consistency (HC) score ${s_{HC}}$ to further assess the part-wise semantic layering ability. 

% \begin{figure}
% \centering
% \includegraphics[width=1\linewidth]{sec/pic/hi_com.pdf}
% \caption{More comparison of hierarchy on the Hi-LERF dataset.}
% \label{viz_hi}
% \end{figure}

\begin{table*}[!htp]
	\centering
 \caption{Comparisons of mIoU and HC scores  between our model and SOTA hierarchy methods on
Hi-LERF and Hi-3DOVS datasets. L1-3: Level 1-3. Avg: Average.} 
	%\resizebox{8cm}{
	\scalebox{0.95}{% Please add the following required packages to your document preamble:
% \usepackage{multirow}
% \usepackage{multirow}\
\small
	\setlength{\tabcolsep}{10pt}
%\begin{tabular}{lcccccccccccc}
\begin{tabular}{lccccc|ccccc}
\hline
\multirow{3}{*}{Method} & \multicolumn{5}{c|}{Hi-LERF}                               & \multicolumn{5}{c}{Hi-3DOVS}                                                      \\ \cline{2-11} 
                        & \multicolumn{4}{c|}{Instance (mIoU)}           & Part (HC) & \multicolumn{4}{c|}{Instance (mIoU)}             & \multicolumn{1}{c}{Part (HC)} \\ \cline{2-11} 
                        & L1   & L2   & L3   & \multicolumn{1}{c|}{Avg.} & Overall   & L1   & L2     & L3   & \multicolumn{1}{c|}{Avg.} & \multicolumn{1}{c}{Overall}   \\ \hline
                        \textbf{\emph{3DGS-based}} &&&&\multicolumn{1}{c|}{}&\multicolumn{1}{c|}{}&&&&\multicolumn{1}{c|}{}\\
OpenGaussian~\cite{wu2024opengaussian}                                & 31.5     & 14.2      & 7.1     &\multicolumn{1}{c|} {17.6}     & 15.1        & 68.5     & 18.4     & 11.3    &\multicolumn{1}{c|}{32.7}   & 19.3          \\
Gaussian Grouping~\cite{ye2024gaussiangroupingsegmentedit} &\cellcolor{top3}35.7 &15.9 & 12.8&\cellcolor{top3}21.5 &17.3 & \cellcolor{top3}72.6& 20.6& 15.5& 36.2&23.8 \\
LangSplat~\cite{qin2024langsplat}                                             & \cellcolor{top2}38.9     & \cellcolor{top2}21.8     &\cellcolor{top3} 13.4     & \multicolumn{1}{c|}{\cellcolor{top2}24.7}    & 18.9           & \cellcolor{top2}74.7     & \cellcolor{top2}35.9     & 17.2    & \multicolumn{1}{c|}{\cellcolor{top2}42.6}   & 25.8          \\ \hline
\textbf{\emph{Hierarchical-based}} &&&&\multicolumn{1}{c|}{}&\multicolumn{1}{c|}{}&&&&\multicolumn{1}{c|}{}\\
VCH~\cite{Kang_2024_CVPRHierarchical}                                                    & 26.3     &16.1     &\cellcolor{top2}15.6    &\multicolumn{1}{c|}{19.3}   & \cellcolor{top2}28.6       &51.1     &26.5     & \cellcolor{top2}25.8    &\multicolumn{1}{c|}{34.5}   &\cellcolor{top2}38.4        \\
OmniSeg3D-GS~\cite{ying2024omniseg3d}                                              & 24.5     & \cellcolor{top3}18.3     & 11.2    & \multicolumn{1}{c|}{18.0}   & \cellcolor{top3}21.5       & 58.3     & \cellcolor{top3}32.1     & \cellcolor{top3}19.3    & \multicolumn{1}{c|}{\cellcolor{top3}36.6}    & \cellcolor{top3}29.7          \\ \hline
\textbf{Ours}                                        & \textbf{\cellcolor{top1}45.1}     & \textbf{\cellcolor{top1}38.5}     & \textbf{\cellcolor{top1}33.7}    & \multicolumn{1}{c|}{\textbf{\cellcolor{top1}39.1}}    &\textbf{\cellcolor{top1}56.9}          & \textbf{\cellcolor{top1}86.2}     &\textbf{\cellcolor{top1}59.8}    & \textbf{\cellcolor{top1}38.6}    & \multicolumn{1}{c|}{\textbf{\cellcolor{top1}61.5}}   & \textbf{\cellcolor{top1}65.9}          \\ \hline
\end{tabular}}
\label{hi_com}
\end{table*}

\begin{figure*}[h]
\centering
\includegraphics[width=1\linewidth]{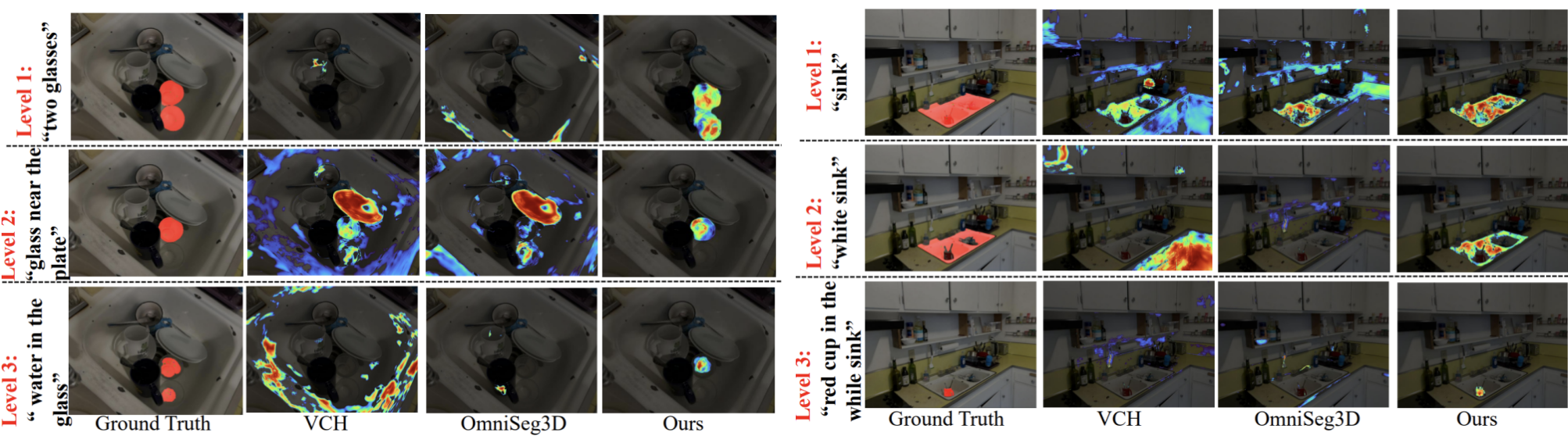}
\caption{Comparison of hierarchy on the Hi-LERF dataset.}
\label{hierarchy dataset}
\end{figure*}

\subsection{Experiments and Results}
Notably, to ensure the fairness of experimental comparisons, it is important to clarify that our evaluation spans 8 diverse datasets~\cite{kerr2023lerf,straub2019replica,3DOVS,dai2017scannet,sun2020scalabilitywaymo}, whereas many existing baselines report results on only a subset of them. Moreover, due to the unavailability of source code or implementation details, several of these methods are not reproducible. As a result, we have made every effort to adopt the officially reported results for each method on the corresponding datasets, as stated in their original papers, to maintain a fair and consistent comparison.

\subsubsection{Comparison on LERF}
%As shown in Table~\ref{lerf} and Fig.~\ref{viz_seg}, our modelachieved the best quantitative results across the SOTA 3D scene open-vocabulary query methods~\cite{qin2024langsplat,Kim_2024_CVPRgrop} on the semantic segmentation and localization tasks, surpassing the CGC~\cite{silva2024contrastivegaussianclusteringweakly} and OpenGaussian models, which also adopt the point-optimized contrastive learning by $1.9$ and $43.8$ mIoU on average. 
%As demonstrated in Table~\ref{lerf} and Fig.~\ref{viz_seg} and \ref{viz_seg2}, our model achieved superior results compared to the SOTA 3D open-vocabulary query methods~\cite{kerr2023lerf,shi2024LEGaussians} in both tasks.
As evidenced by Table~\ref{lerf} and Figures~\ref{viz_seg} and \ref{viz_seg2}, our model consistently outperforms state-of-the-art 3D open-vocabulary querying methods~\cite{kerr2023lerf,shi2024LEGaussians} across both tasks, encompassing approaches based on 2D supervision~\cite{li2022lseg}, NeRF-based representations~\cite{kerr2023lerf,3DOVS,miao2025laser,ying2024omniseg3d,lee2025rethinkingopenvocabularysegmentationradiance}, and 3D Gaussian Splatting frameworks~\cite{qin2024langsplat,shi2024LEGaussians,ying2024omniseg3d,wu2024opengaussian,jun2025dr,szilagyi2025slag,lee2025rethinkingopenvocabularysegmentationradiance,liang2024supergseg,ji2025fastlgs,ye2024gaussiangroupingsegmentedit}.
We reproduced the class-agnostic model OmniSeg3D~\cite{ying2024omniseg3d} under identical experimental settings. Query results were obtained by calculating regions where the similarity between text queries and semantic features exceeded a set threshold.
Compared to OmniSeg3D~\cite{ying2024omniseg3d}, which also explores hierarchical semantic information but relies on 2D foundation models, our approach achieves substantial performance gains across five scenes in both tasks.
It surpasses the OpenGaussian~\cite{wu2024opengaussian}, which also utilizes point-optimization. Our model still exhibits clear superiority compared to Gaussian Grouping~\cite{ye2024gaussiangroupingsegmentedit}, on semantic segemantation task and FastLGS~\cite{ji2025fastlgs} on localization task.
%, by  $21.1$ average mIoU on the LERF dataset.

% \begin{figure}
% \centering
% \includegraphics[width=0.95\linewidth]{sec/pic/seg3d_6.pdf}
%     \caption{Comparison of open-vocabulary semantic query on the ``sofa" scene of 3D-OVS and Hi-3DOVS datasets.}
% \label{viz_loca}
% \end{figure}
Besides, our model outperforms LangSplat~\cite{qin2024langsplat}, which employs hierarchical 2D masks, validating the efficacy of our 3D hierarchical semantic tree.
We also illustrate the qualitative results in Fig.~\ref{viz_seg} and \ref{viz_seg2}. LangSplat~\cite{ying2024omniseg3d} struggles with correct hierarchical semantic instances. OmniSeg3D~\cite{ying2024omniseg3d} has difficulty in challenging queries with complex semantics like ``pirate hat" and Gaussian Grouping fails to segment internal part-wise features, such as ``onion segments in a bowl". 
Our model effectively captures features across different semantic levels.

\begin{figure*}[!htp]
\centering
\includegraphics[width=0.9\linewidth]{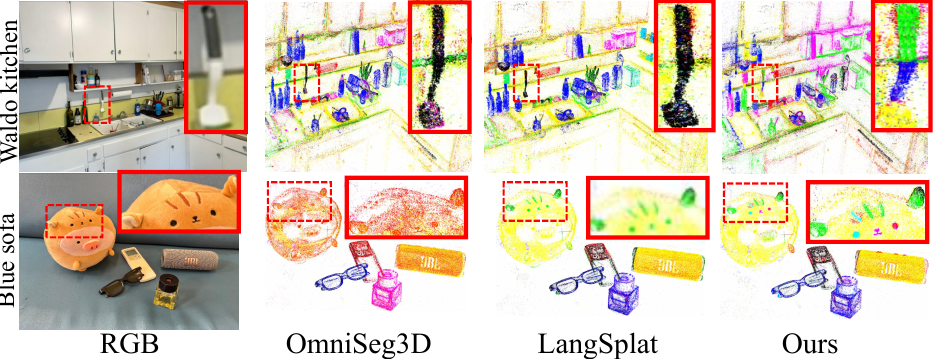}
\caption{Comparison of hierarchy semantic features.}
\label{hierarchy point-optimized}
\end{figure*}
\begin{figure}[!htp]
\centering
\includegraphics[width=1\linewidth]{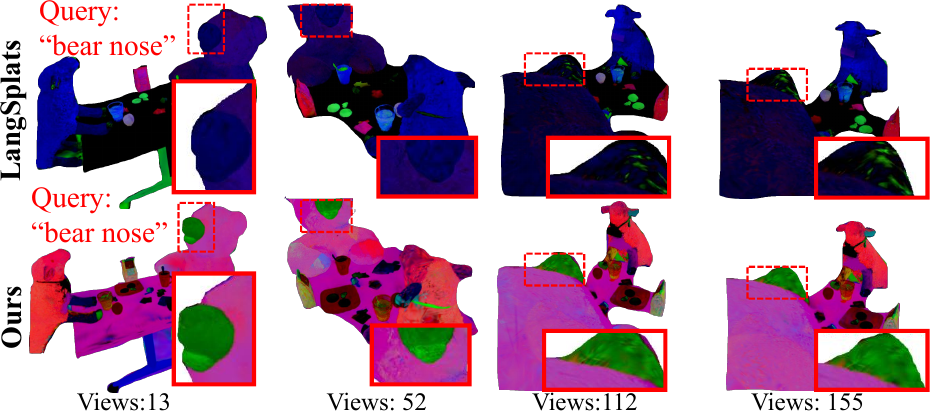}
\caption{Comparison of consist semantics of ``bear nose".}
\label{consist}
\end{figure}

\subsubsection{Comparison on 3D-OVS}
We provide quantitative results for 3D semantic segmentation on the 3D-OVS~\cite{3DOVS} in Table~\ref{3d} and qualitative results in Fig. \ref{viz_loca}. Our model outperforms both 2D-based methods~\cite{li2022lseg,Xu_2023_CVPRODISE,Liang_2023_CVPRseg}, NeRF-based representation~\cite{kerr2023lerf,kobayashi2022decomposingFFD,lee2025rethinkingopenvocabularysegmentationradiance,3DOVS,miao2025laser} and 3DGS-based approaches~\cite{kobayashi2022decomposingFFD,kerr2023lerf,3DOVS,zuo2025fmgs,ji2025fastlgs,ye2024gaussiangroupingsegmentedit} across 5 scenes.
%GaussianGrouping~\cite{ye2024gaussiangroupingsegmentedit} relies on 2D segmentation masks to enforce multi-view consistency, which overlooks blind spots and 3D inconsistencies, thereby diminishing performance. 
Although GaussianGrouping~\cite{ye2024gaussiangroupingsegmentedit} outperforms us in the ``bed" scene, its reliance on 2D segmentation masks to lead to overlook blind spots and 3D inconsistencies, ultimately impairing performance in other scenes.
CGC~\cite{silva2024contrastivegaussianclusteringweakly} and FastLGS~\cite{ji2025fastlgs} overlook the inherent hierarchical semantic structure within feature representations, which hinders their ability to accurately capture deeper-level semantic relationships.
While LangSplat~\cite{qin2024langsplat} incorporates hierarchical semantics, its 2D structure limits its ability to query 3D hierarchical semantics.
%Although LangSplat~\cite{qin2024langsplat} incorporates hierarchical semantics but is constrained by its 2D structure, it is hard to represent 3D hierarchical semantics. 
As shown in Fig.~\ref{viz_loca}, our model precisely segments complex hierarchical semantics.

\subsubsection{More Comparisons on ScanNet, Replica, Waymo } We conducted more experiments on ScanNetv2~\cite{dai2017scannet}, Waymo~\cite{sun2020scalabilitywaymo}, and ScanNet200~\cite{rozenberszki2022languagescannet200} for semantic segmentation task, and Replica~\cite{straub2019replica}, and ScanNet200~\cite{rozenberszki2022languagescannet200} for instance segmentation task, as shown in Table~\ref{replica1}, \ref{scannet3} and \ref{waymo1}. 
We conducted comprehensive comparisons between our model and representative 2D-based~\cite{li2022lseg}, NeRF-based~\cite{kerr2023lerf,3DOVS,ying2024omniseg3d,miao2025laser}, and 3D Gaussian Splatting (3D-GS)-based approaches~\cite{qin2024langsplat,ying2024omniseg3d,shi2024LEGaussians,wu2024opengaussian,jun2025dr,szilagyi2025slag}. As shown in Table~\ref{replica1}, we performed detailed evaluations across eight scenes on the Replica dataset~\cite{straub2019replica}, reporting mIoU and mAcc metrics, where our model consistently achieved the best performance across all scenarios.

For the ScanNet dataset~\cite{dai2017scannet}, we carried out experiments on ScanNetv2~\cite{dai2017scannet} for semantic segmentation under various settings involving 19, 15, and 10 classes, as well as on ScanNet200~\cite{rozenberszki2022languagescannet200} for both semantic and instance segmentation. As shown in Table~\ref{scannet3}, our model demonstrates clear superiority across all configurations, highlighting its strong capability in understanding diverse semantic features in 3D scenes.

Additionally, we evaluated our model on the Waymo dataset~\cite{sun2020scalabilitywaymo} in Table~\ref{waymo1}, where it outperformed all three SOTA baselines~\cite{ye2024gaussiangroupingsegmentedit,wu2024opengaussian,qin2024langsplat}. This further confirms the robustness and generalization ability of our model, even in challenging outdoor environments.

More comparisons demonstrate that our model outperforms others~\cite{qin2024langsplat,wu2024opengaussian} in both indoor and outdoor datasets, showcasing its ability to capture {deeper semantics for real-world applications}.

\begin{table}[h]
	\centering
 \caption{Ablation study. HST: Hierarchical Semantic Tree, Initial-2D: Initial 2D Semantic Levels. 3D-HC: 3D Hierarchical Cluster. CL: Contrastive Learning. } 
	\scalebox{0.85}{
\small
	\setlength{\tabcolsep}{1pt}
\begin{tabular}{cccc|c|c|cc}
\hline
\multicolumn{4}{c|}{Method}                                                                         & LERF(Avg.) & \multicolumn{1}{l|}{3D-OVS(Overall)} & \multicolumn{2}{c}{Hi-LERF(Avg.)} \\ \hline
\multicolumn{2}{c|}{HST}                                     & \multicolumn{2}{c|}{CL}              & Localization  & \multicolumn{1}{c|}{Segmentation}    & \multicolumn{2}{c}{Segmentation}     \\ \hline
\multicolumn{1}{c|}{Initial-2D} & \multicolumn{1}{c|}{3D-HC} & \multicolumn{1}{c|}{Instance} & Part & Accuracy      & mIoU                                 & mIoU              & HC               \\ \hline
×                               & ×                          & ×                             & ×    & 87.4        & 93.4                              & 24.7             & 18.9          \\
\checkmark                               & ×                          & ×                             & ×    &88.0     & 93.8                                & 25.5           & 20.1        \\
\checkmark                                 & \checkmark                            & ×                             & ×    &90.6         & 94.8                                 & 31.8              & 42.2             \\
\checkmark                                 & \checkmark                            &\checkmark                               & ×    & 92.0         & 95.6                                 & 36.3              & 53.8             \\
\checkmark                                & \checkmark                           & ×                             &\checkmark     & 91.5        & 95.2                               & 34.5              & 49.3             \\
\checkmark                                & \checkmark                           & \checkmark                              & \checkmark      & 92.7          & 96.1                                 & 39.1              & 56.9             \\ \hline
\end{tabular}}
\label{ablation}
\end{table}

\subsubsection{Comparison on Hierarchy}
As shown in Table~\ref{hi_com} and Fig.~\ref{hierarchy dataset}, our model outperforms other SOTA hierarchical semantic methods~\cite{he2024viewconsistenthierarchical3dsegmentation,ying2024omniseg3d} on Hi-LERF and Hi-3DOVS, consistently achieving the highest mIoU and hierarchical consistency (HC) scores, especially at higher semantic levels, highlighting its superior hierarchical semantic understanding. Fig~\ref{hierarchy dataset} showcases the superior performance on the hierarchical semantic dataset. 
While VCH~\cite{he2024viewconsistenthierarchical3dsegmentation} is restricted to single-instance segmentation, it struggles with the ``white sink" from a ``white cabinet."
Besides, OmniSeg3D~\cite{ying2024omniseg3d} focuses on coarse semantic distinctions, overlooking internal semantic hierarchies, and struggling with higher-level semantics like a ``red cup in a white sink."
%multiple objects in complex 3D scenes, such as distinguishing between a ``white sink" and a ``white cabinet."
%Besides, while OmniSeg3D~\cite{ying2024omniseg3d} learns hierarchical semantics, it focuses on coarse distinctions between instances, neglecting internal semantic hierarchies. As a result, it struggles with higher-level semantics, like distinguishing a ``red cup in a white sink."
%although OmniSeg3D~\cite{ying2024omniseg3d} also learns hierarchical semantics, it addresses coarse-grained semantic distinctions between instances and neglects the internal semantic hierarchy. Consequently, it performs inadequately in capturing higher-level semantics, such as distinguishing a ``red cup in the white sink."
%it primarily focuses on coarse-grained semantics between instances and overlooks internal semantic hierarchy, resulting in poor performance in higher-level semantics, such as a ``red cup in the white sink." 
In contrast, 
%our model not only effectively segments instances with varying semantic similarity in complex scenes but also excels in capturing higher-level semantic distinctions.
as Fig.~\ref{hierarchy point-optimized} shows, 
our model adeptly learns hierarchy features across various semantic levels, such as the ``head, middle, and tail of a spatula," where other methods struggle. %More comparisons of the hierarchy are in Appendix E.

\subsubsection{Comparison on Consistency} 
We further analyzed the model's ability to capture consistent semantic features. As shown in Fig.~\ref{consist}, we selected 4 different views of the ``teatime" scene from the LERF dataset to compare semantic features of the ``bear nose." It shows that our model effectively learns view-consistent semantic features of the bear's nose across views, while the SOTA model, LangSplat~\cite{qin2024langsplat}, fails to capture complete and view-consistent features.

%While our model has slightly longer training times due to hierarchical semantic learning, its inference speeds are competitive with others, making it \textbf{cost-efficient}.

%Fig. 5 provides a detailed illustration of our model's superior performance in multi-level semantic segmentation. Although OmniSeg3D also learns hierarchical semantics, it primarily focuses on coarse-grained instance-wise segmentation, making it less effective at understanding higher-level semantics, such as the ``logo on a bag." Besides, while VCH aims to generate consistent 3D hierarchical semantic structures, it remains constrained to single-instance segmentation and struggles with multi-layer semantic segmentation in complex 3D scenes involving multiple objects, such as a ``bear's leg" or a ``tag on a sheep's neck." In contrast, our model not only effectively segments instances of varying semantic similarity but also excels in segmenting part-wise hierarchical semantics within objects.

\begin{table*}[!htp]
	\centering
    \small
 \caption{Computational costs. s/q is the average time per query.} 
      \setlength{\tabcolsep}{15pt}
	\scalebox{0.95}{% Please add the following required packages to your document preamble:
% \usepackage{multirow}
% \usepackage{multirow}\、
\small
\begin{tabular}{lcccccc}
\hline
\multicolumn{1}{c}{\multirow{2}{*}{Method}} & \multicolumn{2}{c}{Train (LERF)} & \multicolumn{4}{c}{Inference Speed (s/q)}                                                                                   \\ \cline{2-7} 
\multicolumn{1}{c}{}                        & Time (min)      & Cost (GB)      & \multicolumn{1}{c}{Figurines} & \multicolumn{1}{c}{Ramen} & \multicolumn{1}{c}{Teatime} & \multicolumn{1}{c}{Waldo kitchen} \\ \hline

OpenGaussian~\cite{wu2024opengaussian}                                & 50                                     & \multicolumn{1}{c|}{20}                                   & 0.37      & 0.35  & 0.33    & 0.36          \\
VCH~\cite{ying2024omniseg3d}                                & 71                                  & \multicolumn{1}{c|}{24}                                   & 0.39     & 0.36 & 0.30   & 0.37         \\
Gaussian Grouping~\cite{ye2024gaussiangroupingsegmentedit}                                & 82                                & \multicolumn{1}{c|}{24}                                   & 0.31     & 0.29 & 0.27   & 0.28        \\
OmniSeg3D-GS~\cite{ying2024omniseg3d}                                & 105                                   & \multicolumn{1}{c|}{24}                                   & 0.33     & 0.31 & 0.34   & 0.32         \\
LangSplat~\cite{qin2024langsplat}                                  & {25}x3                            & \multicolumn{1}{c|}{4}                           & 0.28      & 0.26  & 0.23    & {0.25}         \\ \hline
\textbf{Ours}                                        & 114                                    & \multicolumn{1}{c|}{24}                                   & \multicolumn{1}{c}{\textbf{0.25}}      & \textbf{0.24}  & \textbf{0.20}    & \textbf{0.23}         \\ \hline
\end{tabular}
}
%\setlength{\abovecaptionskip}{-0.00cm}
%\captionsetup{font={small}}
	\label{efficiency2}
\end{table*}
\begin{table*}[!htp]
	\centering
    \small
 \caption{Computational costs after samples pruning strategy .} 
      \setlength{\tabcolsep}{10pt}
	\scalebox{0.95}{% Please add the following required packages to your document preamble:
% \usepackage{multirow}
% \usepackage{multirow}\、
\small
\begin{tabular}{ccllcllcc}
\hline
\multirow{3}{*}{Method}         & \multicolumn{3}{c|}{LERF}                                                                          & \multicolumn{3}{c|}{3D-OVS}                                                                        & \multicolumn{2}{c}{Hi-LERF(Average)} \\ \cline{2-9} 
                                & \multicolumn{2}{c|}{\begin{tabular}[c]{@{}c@{}}Train(average)\end{tabular}} & \multicolumn{1}{c|}{Inference}     & \multicolumn{2}{c|}{\begin{tabular}[c]{@{}c@{}}Train(average)\end{tabular}} & \multicolumn{1}{c|}{Inference}     & \multicolumn{2}{c}{Segmentation}     \\ \cline{2-9} 
                                & time(min)                                   & \multicolumn{1}{c|}{cost(GB)}        & \multicolumn{1}{c|}{Speed (s/q)}   & time(min)                                   & \multicolumn{1}{c|}{cost(GB)}        & \multicolumn{1}{c|}{Speed (s/q)}   & mIoU              & HC               \\ \hline
\multicolumn{1}{l}{original}    & \multicolumn{1}{l}{114}                     & \multicolumn{1}{c|}{24}                                  & \multicolumn{1}{c|}{0.24}         & \multicolumn{1}{l}{121}                     & \multicolumn{1}{c|}{24}                                  & \multicolumn{1}{c|}{0.21}          & 39.1              & 56.9             \\
\multicolumn{1}{l}{pruning} & \multicolumn{1}{l}{\textbf{46({-68})}}        & \multicolumn{1}{c|}{\textbf{8({-16})}}                         & \multicolumn{1}{c|}{\textbf{0.21}} & \multicolumn{1}{l}{\textbf{52({-69})}}        & \multicolumn{1}{c|}{\textbf{8({-16})}}                          & \multicolumn{1}{c|}{\textbf{0.17}} & 38.8              & 56.4             \\ \hline
\end{tabular}

}
%\setlength{\abovecaptionskip}{-0.00cm}
%\captionsetup{font={small}}
	\label{reduce}
\end{table*}

\begin{table}[h]
 % \vspace{-0.42cm}
	\centering
 \caption{Influence of different coverage threshold  $\theta$.} 
      \setlength{\tabcolsep}{2pt}
	\scalebox{1}{% Please add the following required packages to your document preamble:
% \usepackage{multirow}
% \usepackage{multirow}\、
\small
\begin{tabular}{cccc}
\hline
\multirow{3}{*}{Coverage threshold  $\theta$} & 3D-OVS(Overall) & \multicolumn{2}{c}{Hi-LERF(Average)} \\ \cline{2-4} 
                                    & Segmentation    & \multicolumn{2}{c}{Segmentation}     \\ \cline{2-4} 
                                    & mIoU            & mIoU              & HC               \\ \hline
0.6                                 & 95.4            & 38.4              & 56.3             \\
0.7                                 & 95.5            & 38.4              & 56.2             \\
0.8                                 & 95.8            & 38.7              & 56.5             \\
0.9                                 & 96.1            & 39.1              & 56.9             \\ \hline
\end{tabular}}
%\setlength{\abovecaptionskip}{-0.00cm}
%\captionsetup{font={small}}
	\label{threshod}
\end{table}

\subsubsection{Comparison on Efficiency}
As shown in Table~\ref{efficiency} and \ref{efficiency2}, we compare the average training time for one scene and the memory costs with other SOTA models under identical settings. The results indicate that although our model requires more training time due to hierarchical semantic learning, the memory costs are comparable to VCH~\cite{he2024viewconsistenthierarchical3dsegmentation} and OmniSeg3D~\cite{ying2024omniseg3d}, yet our model achieves superior hierarchical semantic results. Besides, we adapt contrastive sample pruning to reduce cost, removing redundant pairs based on feature similarity, retaining the most distinct samples in contrastive loss.
{Table~\ref{reduce}} shows the improvements, reduce costs by {nearly 1/3 while preserving performance}.
\begin{table}
 % \vspace{-0.42cm}
	\centering
 \caption{Efficiency comparison. -GS: based on Gaussian Splatting. ``$\times 3$": LangSplat~\cite{qin2024langsplat} is trained separately on 3 different semantic levels, and the best result among the three is selected.} 
      \setlength{\tabcolsep}{1pt}
	\scalebox{0.9}{% Please add the following required packages to your document preamble:
% \usepackage{multirow}
% \usepackage{multirow}\、
\small
\begin{tabular}{lcccc}
\hline
\multicolumn{1}{c}{\multirow{3}{*}{Method}} & \multicolumn{2}{c}{Hi-LERF(Average)}              & \multicolumn{1}{c}{\multirow{2}{*}{Training Time}} & \multicolumn{1}{c}{\multirow{2}{*}{Memory Cost}} \\ \cline{2-3}
\multicolumn{1}{c}{}                        & \multicolumn{2}{c}{Segmentation}                  & \multicolumn{1}{c}{}                            & \multicolumn{1}{c}{}                      \\ \cline{2-5} 
\multicolumn{1}{c}{}                        & \multicolumn{1}{c}{mIoU} & \multicolumn{1}{c}{HC} & min                                             & GB                                        \\ \hline
OpenGaussian~\cite{wu2024opengaussian}                                & 17.6                     & 15.1                   & 50                                              & 20                                        \\
LangSplat~\cite{qin2024langsplat}                                   & 24.7                     & 18.9                   & \textbf{25} $\times 3 $                                    & \textbf{4}                                \\
VCH~\cite{he2024viewconsistenthierarchical3dsegmentation}                                         & 19.3                     & 28.6                   & 71                                              & 24                                        \\
Gaussian Grouping~\cite{ye2024gaussiangroupingsegmentedit}                                         & 21.5                   &17.3              & 82                                              & 24                                        \\
OmniSeg3D-GS~\cite{ying2024omniseg3d}                                   & 18.0                       & 21.5                   & 105                                              & 24                                        \\ \hline
\textbf{Ours}                                        & \textbf{39.1}            & \textbf{56.9}          & 114                                             & 24                                        \\ \hline
\end{tabular}}
%\setlength{\abovecaptionskip}{-0.00cm}
%\captionsetup{font={small}}
	\label{efficiency}
\end{table}
\subsection{Ablation Study}
As Table~\ref{ablation} shows, We conduct ablation studies to validate the efficacy of proposed methods. 
We take LangSplat~\cite{qin2024langsplat} as the baseline, as shown in row 1.
\subsubsection{3D Hierarchical Cluster}
The comparison between rows 2-3 in Table ~\ref{ablation} reveals that using 3D point-level instance clustering significantly improves semantic segmentation by $22.1$ HC score on Hi-LERF and $1.0$ mIoU on 3D-OVS. 
This shows that 3D point-level hierarchical clustering addresses multi-view inconsistencies in 2D models, proving its effectiveness in capturing 3D consistent hierarchical semantics.
\begin{table}[h]
 % \vspace{-0.42cm}
	\centering
 \caption{Influence of different similarity degrees $\Omega$.} 
      \setlength{\tabcolsep}{2pt}
	\scalebox{1}{% Please add the following required packages to your document preamble:
% \usepackage{multirow}
% \usepackage{multirow}\、
\small
\begin{tabular}{cccc}
\hline
\multirow{3}{*}{Similarity degrees $\Omega$} & LERF(Average) & \multicolumn{2}{c}{Hi-LERF(Average)} \\ \cline{2-4} 
                                             & Segmentation  & \multicolumn{2}{c}{Segmentation}     \\ \cline{2-4} 
                                             & mIoU          & mIoU              & HC               \\ \hline
2                                            & 66.3         & 36.4              & 53.3             \\
\textbf{10}                                  & \textbf{68.4} & \textbf{39.1}     & \textbf{56.9}    \\
100                                          & 68.0          & 37.6              & 55.8             \\
1000                                         & 65.2          & 35.3              & 52.7             \\ \hline
\end{tabular}}
%\setlength{\abovecaptionskip}{-0.00cm}
%\captionsetup{font={small}}
	\label{omega}
\end{table}
\subsubsection{3D Hierarchical Semantic Tree}
%As shown in rows 1-3 of Table ~\ref{ablation}, the model equipped with a 3D hierarchical semantic tree which is composed of initial semantic levels and 3D hierarchical cluster further improves semantic segmentation by $23.5$ mIoU and enhances localization tasks by $11.6\%$ accuracy. 
As shown in rows 1-3 of Table ~\ref{ablation},  the 3D hierarchical semantic tree which consists of initial semantic levels and 3D hierarchical cluster, improves semantic segmentation by $17.1$ mIoU on Hi-LERF and $3.2\%$ localization accuracy on LERF.
It benefits from the 3D hierarchical semantic tree which effectively captures layered semantics in complex 3D scenes and differentiates between varying semantic similarities, such as ``stuffed bear" and ``bear nose."  
%More ablations are in Appendix C.
\subsubsection{Contrastive Learning}
From rows 4 and 5 in Table ~\ref{ablation}, we observe that the instance-wise and part-wise contrastive losses improves the semantic segmentation by $4.5\%$ and $2.7\%$ mIoU on Hi-LERF. 
It demonstrates that instance-wise contrastive loss captures multi-level semantic features and improves the differentiation of similar semantics.
%It demonstrates that instance-wise contrastive loss not only effectively captures different semantic-level features  but also enhances the differentiation of similarities between semantic features. %(e.g., the coffee cup and the water inside the cup). 
Besides, the part-wise contrastive loss improves discrimination between similar semantic features (e.g., the bear's nose and mouth). While distinguishing similar features is challenging, our approach captures both external and internal hierarchies, offering a deeper 3D scene understanding.

\subsection{Parameter Discussions}
\subsubsection{Coverage threshold  $\theta$}

As shown in Table~\ref{threshod}, we discuss the impact of varying cover thresholds on the performance of our open-vocabulary query semantic segmentation task across the 3D-OVS and Hi-LERF datasets. 
The best results are obtained when the coverage threshold is set to $0.9$.
The results in the table indicate that a higher cover threshold implies stricter coverage requirements across the three semantic levels, suggesting that more rigorous semantic layering enhances the model's ability to differentiate between hierarchical semantics.

\subsubsection{Similarity degrees $\Omega$}

In our proposed instance-wise loss, we aim to approximate the ratio between the distances of sample pairs to be equal to the ratio of their similarity degrees $\Omega$. Consequently, appropriately assigning similarity distances to pairs guided by their similarity is crucial in our method. In this ablation study, we explore the impact of different values of $\Omega$ (i.e., preset similarity bases) on model performance. For instance, the hyperparameter $\Omega$ ranges from 2 to 1000, as depicted in Table~\ref{omega}. We observe that our method achieves optimal performance when $\Omega$ = 10. The results indicate that, on the LERF dataset, the distance ratio between adjacent similar sample pairs approaches 10.
 If the hyperparameter $\Omega$ is set too small (e.g. 2) or too large (e.g. 1000), it will lead to incorrect distance ratios between sample pairs, thereby impairing the proper semantic hierarchical relationships between semantic features.

\section{Conclusion}
In this paper, we propose Hi-LSplat,  a view-consistent 3D hierarchical  Language Gaussian field for 3D open-vocabulary query. 
The innovation lies in using a 3D hierarchical semantic tree to capture 3D view-dependent semantics, coupled with instance-wise and part-wise contrastive learning to grasp complex hierarchies. 
We also created two datasets to better evaluate hierarchical semantics. Hi-LSplat excels in 8 datasets. Codes and datasets will be released.

\noindent \textbf{Limitations.} 1) Free-form semantic querying of 3D scenes remains challenging. We plan to extend open-vocabulary queries to free-form queries without any training priors. 2) The cluster and contrastive learning slightly increase training time and resource but remain cost-efficient.

%\begin{thebibliography}{1}
\bibliographystyle{IEEEtran}

\bibliography{ref.bib}

\vfill

\end{document}